\documentclass[conference]{IEEEtran}

\usepackage{cite}
\usepackage{amsmath,amssymb,amsfonts}
\usepackage{algorithmic}
\usepackage{graphicx}
\usepackage{textcomp}
\usepackage{xcolor}
\def\BibTeX{{\rm B\kern-.05em{\sc i\kern-.025em b}\kern-.08em
    T\kern-.1667em\lower.7ex\hbox{E}\kern-.125emX}}

\usepackage{multirow}
\usepackage{accents}
\usepackage{mathbbol, mathrsfs}
\usepackage[ruled, vlined, linesnumbered]{algorithm2e}
\usepackage{subcaption}
\usepackage{enumitem}
\usepackage[font=small]{caption}
\usepackage{booktabs}
\usepackage{hyperref}

\DeclareMathOperator{\Beta}{Beta}
\DeclareMathOperator{\AdaIN}{AdaIN}
\DeclareMathOperator{\MixStyle}{MixStyle}
\DeclareMathOperator{\DSU}{DSU}
\DeclareMathOperator{\StyleExplore}{StyleExplore}
\DeclareMathOperator{\StyleShift}{StyleShift}
\newlength{\dhatheight}

\newtheorem{theorem}{Theorem}[section]
\newtheorem{assumption}[theorem]{Assumption}
\newtheorem{remark}[theorem]{Remark}
\newtheorem{lemma}[theorem]{Lemma}
\newtheorem{proposition}[theorem]{Proposition}
\newtheorem{proof}[theorem]{Proof}

\begin{document}

\title{Decentralized Domain Generalization with Style Sharing: Formal Model and Convergence Analysis}

\author{\IEEEauthorblockN{Shahryar Zehtabi$^a$, Dong-Jun Han$^b$, Seyyedali Hosseinalipour$^c$, Christopher G. Brinton$^a$}
\IEEEauthorblockA{$^a$Purdue University, $^b$Yonsei University, $^c$University at Buffalo --- SUNY}
\IEEEauthorblockA{$^a$\{szehtabi,cgb\}@purdue.edu, $^b$djh@yonsei.ac.kr, $^c$alipour@buffalo.edu}}


\maketitle

\begin{abstract}
    Much of federated learning (FL) focuses on settings where the local dataset statistics remain the same between training and testing. 
    However, this assumption often does not hold in practice due to distribution shifts, motivating the development of domain generalization (DG) approaches that leverage source domain data to train models capable of generalizing to unseen target domains. In this paper, we are motivated by two major gaps in existing work on FL and DG: (1) the lack of formal mathematical analysis of DG objectives; and (2) DG research in FL being limited to the star-topology architecture. We develop Decentralized Federated Domain Generalization with Style Sharing (\textsc{StyleDDG}), a decentralized DG algorithm that allows devices in a peer-to-peer network to achieve DG based on sharing style information inferred from their datasets. Additionally, we provide the first systematic approach to analyzing style-based DG training in decentralized networks. We cast existing centralized DG algorithms within our framework and employ their formalisms to model \textsc{StyleDDG}. We then obtain analytical conditions under which the convergence of \textsc{StyleDDG} can be guaranteed. Through experiments on popular DG datasets, we demonstrate that \textsc{StyleDDG} can obtain significant improvements in accuracy across target domains with minimal communication overhead compared to baselines. 
\end{abstract}


\section{Introduction} \label{sec:intro}

\begin{figure*}[t]
    \centering
    \includegraphics[width=0.95\linewidth]{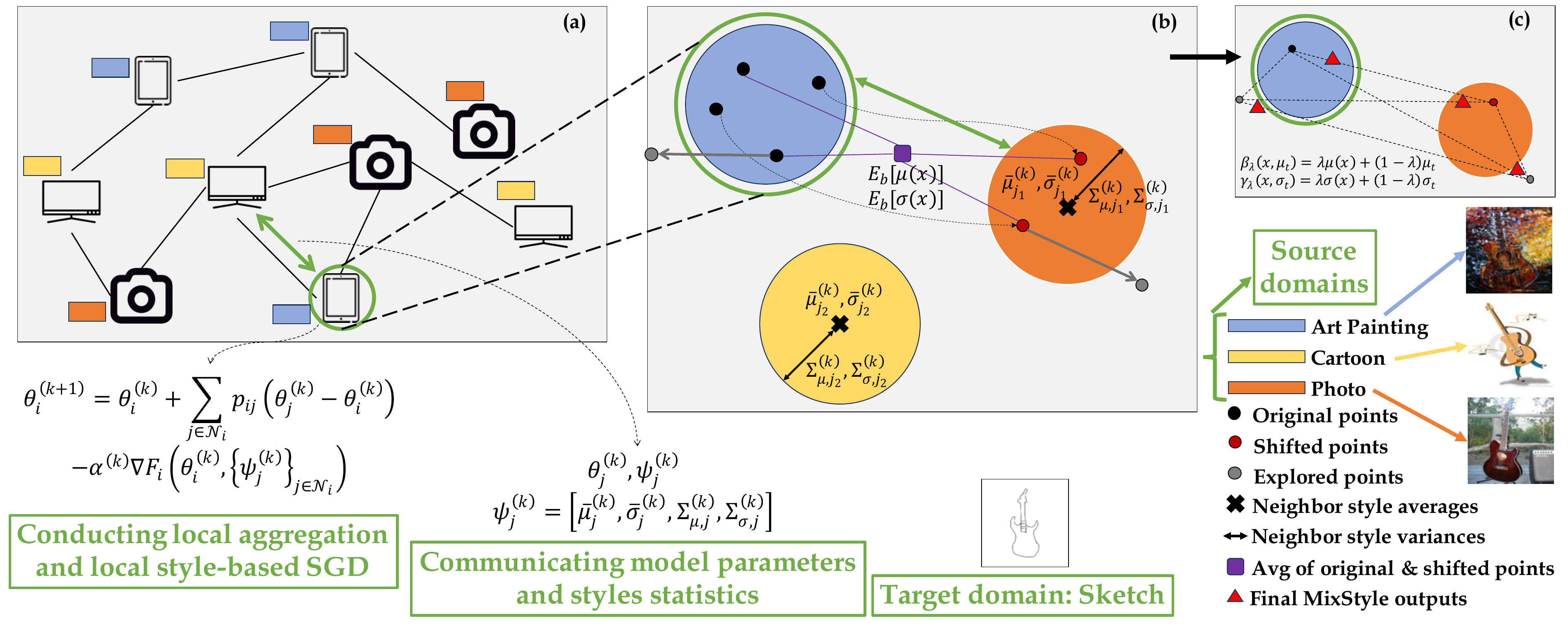}
        \vspace{-2mm}
    \caption{Proposed DDG methodology. (a) A decentralized network comprised of $m=9$ devices, with each device having data from one of three domains (painting, cartoon, photo). The performance of DDG is evaluated on the fourth domain (sketch). (b) In \textsc{StyleDDG}, each device first uses the style statistics of a neighbor for style shifting (getting to red dots from the black ones) for a portion its batch size. Then it concatenates the original plus the shifted points and chooses a new portion of them to do style extrapolation with and generate new styles (grey dots). The circles shown illustrate the style space at the output of a neural network layer. (c) After randomly shifting a portion of the points in the original mini-batch to a new style, and then randomly extrapolating a new portion of the outputs, we are left with a new set of styles with size equal to the original mini-batch. In \textsc{StyleDDG}, we use these new points (red triangles) and apply $\MixStyle$ within them.}
    \label{fig:sysdiag}
    \vspace{-4mm}
\end{figure*}
Federated learning (FL) enables multiple edge devices to collaboratively train a machine learning (ML) model using data stored locally on each device \cite{kairouz2021advances}. Existing conventional ``star-topology'' FL methods rely on a central server to aggregate the models sent from the clients \cite{mcmahan2017communication}. However, in decentralized networks, a central server may not always be available to perform model aggregations \cite{brinton2025key}.
Motivated by this, recent research has investigated decentralized federated learning (DFL), in which edge devices iteratively communicate their model parameters with their one-hop neighbors after completing their local model training \cite{yuan2024decentralized}. Although synchronization among edge devices that participate in conventional FL is readily guaranteed as they overwrite their local models with the one receiver from the server at the end of each training round, DFL algorithms must incorporate a cooperative consensus strategy into their learning process so that devices gradually learn the same model \cite{nedic2009distributed}.

When training ML models using FL/DFL, it is typically assumed that test data are drawn from the same probability distribution as training data \cite{zhou2022domain}. However, this assumption does not always hold; in many real-world applications, a distribution shift can occur between the training set (source) and the test set (target) \cite{ben2010theory}. For example, consider a network of autonomous vehicles \cite{filos2020can} with the aim of improving their image classification models using DFL. As these devices are driven mainly during the morning and afternoon, they mostly contain images in daylight (source domains), while their goal is to train a generalized model capable of classifying images taken at night (target domain) as well. ML algorithms have been shown to drastically fail in achieving generalization to such unseen target domains without making special adjustments to their training processes \cite{wang2022generalizing}. Domain generalization (DG) aims to mitigate this issue, with the goal of preventing performance degradation under such distribution shifts \cite{zhou2022domain}.

Due to the practical importance of enabling DG in FL settings, recent research has focused on federated domain generalization (FDG) \cite{baibenchmarking}, where DG algorithms are designed to operate in a distributed manner across edge devices. FDG introduces additional challenges compared to centralized DG, mainly due to the limited number of source domains accessible to each device \cite{li2023federated}. In FDG, in addition to label heterogeneity between devices, source heterogeneity must also be considered, further complicating the goal of achieving DG \cite{wang2024multi}. For instance, in the previously mentioned example, one device may contain images collected in the morning, while another holds images taken in the afternoon, yet the constructed global model is expected to make reliable predictions on night-time images.

\textbf{Research objectives.} A large class of DG algorithms are \textit{style-based}, with style statistics estimated from the source dataset and explored for generalization to the target. These works, whether for centralized training \cite{zhou2022domain} or for star-topology FL \cite{baibenchmarking}, lack a concrete mathematical formulation of the objective function to be optimized in training domain-invariant models. Furthermore, the studied DG algorithms in FL are constrained by their star-topology networks, and cannot be readily extended to arbitrary decentralized networks. In this paper, we present one of the first style-based decentralized FDG algorithms, called \textit{Decentralized Federated Domain Generalization with Style Sharing} (\textsc{StyleDDG}), in which edge devices communicate style information of their local datasets with their one-hop neighbors so that collectively they can train ML models that generalize well to unseen domains. Our \textsc{StyleDDG} algorithm, illustrated in Fig.~\ref{fig:sysdiag}, is inspired by the work in \cite{park2024stablefdg}, where the authors present an FDG algorithm called StableFDG. Unlike \cite{park2024stablefdg}, \textsc{StyleDDG} is a fully decentralized DG algorithm and does not add any additional oversampling or attention layers as in StableFDG. The computational overhead of StableFDG is at least doubled compared to its analogous models, due to its oversampling step and the added attention layer, and this in turn obstructs our understanding of its style exploration strategy in achieving DG performance. Lastly, our other main contribution is to provide the first formal modeling of style-based DG training, both for well-known DG methods and for \textsc{StyleDDG}.

\textbf{Contributions.} Our key contributions are as follows:
\begin{itemize}[leftmargin=5mm, topsep=-1pt]
    \item We present \textsc{StyleDDG}, one of the first fully decentralized domain generalization (DDG) algorithms. In our setup, devices in a decentralized system aim to achieve DG without having any information about the whole network graph, and can only communicate with their one-hop neighbors, unlike in star-topology FL. Specifically, each device in DDG trains an ML model locally using its local dataset, and then it exchanges (i) its ML model parameters and (ii) style statistics of its current batch with its neighbors (see Fig.~\ref{fig:sysdiag}-(a)). Afterward, each device employs style statistics received from its neighbors to explore the style space more effectively for DG (see Fig.~\ref{fig:sysdiag}-(b)).

    \item Among all style-based DG methods, for the first time in the literature, we provide a systematic approach to mathematically formulate the training objective function and its optimization procedure. As the operations of style-based methods have commonalities, we first derive a set of concrete formulations for well-known style-based DG methods. These generalized formulations lay the groundwork for formal modeling of future methods in this emerging field, applicable to both centralized and distributed/federated network settings.
    
    \item 
    Exploiting the analytical framework mentioned above, we present a formal modeling of the objective function of \textsc{StyleDDG} and carry out its convergence analysis that enables us to provide one of the first convergence guarantees among all style-based DG work. Our results illustrate that under certain conditions on the ML model being employed, the smoothness of local loss functions can be guaranteed for each device. Consequently, this ensures that all devices collectively converge to the same optimal model.

    \item Through experiments on two DG datasets, we demonstrate that our \textsc{StyleDDG} method outperforms other centralized/federated DG methods
    applied in a decentralized learning setting, in terms of the final achievable test accuracy on unseen target domains.
\end{itemize}

\section{Related Work} \label{sec:related}

\textbf{Decentralized federated learning (DFL).}
A recent line of work has considered FL over non-star topologies. This includes semi-decentralized FL \cite{parasnis2023connectivity, chen2024taming}, which groups devices into several subnetworks, with devices conducting intra-subnet communications before exchanging their model parameters with the centralized server. Fog learning generalizes this to multi-layer hierarchies between edge devices and backbone servers \cite{hosseinalipour2020federated}. DFL is at the extreme end of this spectrum, where communication is entirely peer-to-peer, with the goal of iteratively updating and exchanging models on the network graph to arrive at an optimal ML model \cite{liu2022general, zhang2022net, zhang2023novel, huang2024overlay}.
Nevertheless, currently there does not exist any DFL algorithm which is tailored for DG. In this paper, we propose a DG methodology tailored to the DFL network setting, leading to one of the first style-based DDG algorithms in the literature.

\textbf{Domain generalization (DG).}
DG aims to achieve generalization to out-of-distribution data from unseen target domains using distinct source domain data for training (see \cite{zhou2022domain, wang2022generalizing} for recent surveys). Although various classes of ML methods can be considered DG -- like meta-learning \cite{li2018learning, balaji2018metareg}, self-supervised learning \cite{carlucci2019domain}, regularization \cite{huang2020self} and domain alignment \cite{ghifary2015domain, li2018domain} -- our focus in this paper is on the family of \textit{feature augmentation, style-based} methods for DG. In particular, algorithms such as $\MixStyle$ \cite{zhou2021domain} and $\DSU$ \cite{li2022uncertainty} achieve DG using feature-level augmentation, where image styles are explored beyond the distribution of styles that exist within the source domains. In the core of these algorithms, they employ $\AdaIN$ \cite{huang2017arbitrary} (adaptive instance normalization) to explore new features/styles during training. Compared to other DG methods, an advantage of style-based methods is that they do not require the domain labels to be known \cite{park2024stablefdg}.

Despite the notable progress in this domain, a major drawback of current research on DG is the lack of concrete formulations for objective functions. Thus, convergence guarantees for these methods have remained elusive. We fill this gap by rigorously formulating the optimization procedures of previous well-known DG methods and our approach, and by providing a convergence analysis in decentralized network settings.


\textbf{Federated domain generalization (FDG).}
Data heterogeneity is a core challenge in FL, as the distribution of data among devices is non-IID \cite{li2020federated}. This challenge becomes more prominent when the distribution of test data is different from the distribution of training data on devices, motivating the development of FDG algorithms.
In this direction, FedBN \cite{lifedbn} proposed not to average the local batch normalization layers to alleviate the problem of feature shift.
In StableFDG \cite{park2024stablefdg}, each device receives the style of one other participating device from the server, and uses it to extrapolate the region outside the existing style domains to achieve generalization.

However, existing FDG methods \cite{liu2021feddg, wu2021collaborative, chen2023federated, zhang2023federated} are tailored for conventional star-topology FL. DFL introduces novel challenges due to the need to incorporate peer-to-peer consensus mechanisms. In this paper, we present one of the first decentralized domain generalization (DDG) algorithms. Furthermore, we are unaware of any style-based DG algorithm with concrete convergence analysis. Our work thus fills an important gap through formal modeling of DG objective functions, which not only yields insight into the theoretical implications of our proposed methodology, but also lays the foundation for rigorous formulation and convergence analysis of future style-based DG methods \cite{kang2022style, park2023test}.

\section{Style-Based Domain Generalization} \label{sec:method}

We begin with our preliminaries in Sec.~\ref{ssec:prelims}, and discuss the problem formulation in Sec.~\ref{ssec:problem}. Finally, we provide a novel mathematical formulation for the objective functions of two centralized DG algorithms in Sec.~\ref{ssec:cen_model}, which we later use in Sec.~\ref{sec:decen_dg} to formulate our \textsc{StyleDDG} algorithm.

\subsection{Preliminaries} \label{ssec:prelims}

\begin{figure*}
    \centering
\includegraphics[width=0.9\linewidth]{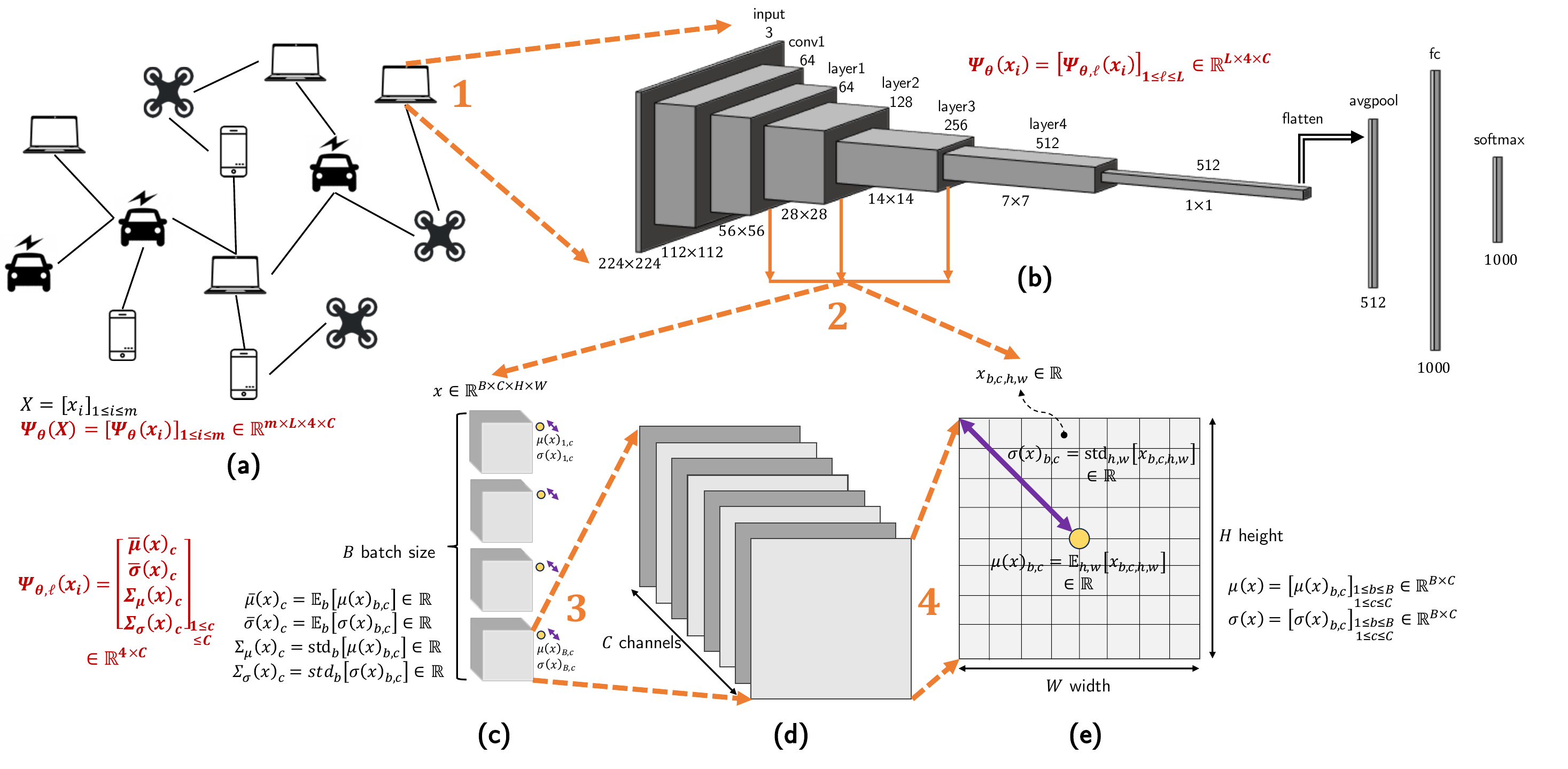}
    \vspace{-2mm}
    \caption{Illustration of style statistics. (a) A decentralized network with devices having access to data from only a single domain. (b) An overview of the ResNet18 model \cite{he2016deep} as an example model at each device, where the number of channels and output size of each block are indicated. We visualize the blocks/layers that \textsc{StyleDDG} is applied to, which correspond to layers where we extract style statistics from. (c) Illustrates the outputs of a particular layer for all instances in the batch of size $B$. The style statistics that each device shares with its neighbors are obtained in the batch-level as shown. (d) For a given instance in the batch, the style statistics are calculated for each channel separately. (e) The $\mu$ and $\sigma$ of style statistics are computed based on the mean and standard deviation of the layer outputs.}
    \label{fig:stylestatistics}
    \vspace{-3mm}
\end{figure*}

\subsubsection{Notation}
We denote the model parameters of a neural network with dimension $n \in \mathbb{Z}^+$ by $\theta \in \mathbb{R}^n$. Let us partition the model $\theta$ into $L \in \mathbb{Z}^+$ parts. We use the notation $\theta_\ell \subset \theta$ with $\ell=1,...,L$ as sub-model with the model parameters of partition $\ell$ of the model $\theta$. Thus, we will have $\theta = [\theta_1^T, ..., \theta_L^T]^T$. We focus much of our presentation on convolutional neural networks (CNNs), given their widespread use in DG methods over vision/image applications. As an example, Fig.~\ref{fig:stylestatistics}-(b) shows how a ResNet model \cite{he2016deep} can be partitioned by its convolutional blocks.

We abstract the operations applied at the end of each layer $\ell$ as $h_{\theta_\ell}(\cdot)$, giving the final CNN output for an input $x$ as $(h_{\theta_L} \circ \cdots \circ h_{\theta_1})(x)$, where $\circ$ denotes the composition operator. Let us denote the loss function as $\mathscr{L}(h_\theta(x), y)$ which takes the predicted outputs $h_\theta(x)$ and the true labels $y$ and calculates the loss based on them.
Also, letting $\tilde{\mathcal{P}}(\cdot)$ denote the power set of a given set, we define a function $\mathcal{P}$ that takes a positive integer input $L \in \mathbb{Z}^+$ and outputs $\mathcal{P}(L) = \tilde{\mathcal{P}}(\{ \ell : 1 \le \ell \le L \})$ with cardinality $|\mathcal{P}(L)| = 2^L$. Finally, let $\mathfrak{S}_\mathcal{I} = \mbox{Permute}(\mathcal{I})$ be the set of all possible permutations of the set $\mathcal{I}$.

\subsubsection{$\AdaIN$}

Adaptive instance normalization ($\AdaIN$) \cite{huang2017arbitrary} plays a key role in style-based  DG methods by enabling style transfers. This technique removes the style information from a data sample and replaces it with the style of another sample. Specifically, the $\AdaIN$ process can be written as follows:
\begin{equation} \label{eqn:adain}
    \AdaIN(x, \mu_t, \sigma_t) = \sigma_t \cdot \frac{x - \mu(x)}{\sigma(x)} + \mu_t,
\end{equation}
where the tensor $x \in \mathbb{R}^{B \times C \times H \times W}$ is taken to be a batch of size $B$ in a specific CNN layer with a channel dimension of $C$, width $W$ and height $H$, and $x_{b,c,h,w}$ denotes the value in the indexed dimensions. Furthermore, $\mu_t, \sigma_t \in \mathbb{R}^{B \times C}$ are the statistics of a target style to which we want to shift the style of $x$ to, and the instance-level mean and standard deviation $\mu(x), \sigma(x) \in \mathbb{R}^{B \times C}$ for each channel are calculated as
\vspace{-2mm}
\begin{equation} \label{eqn:mu_sig}
    \begin{gathered}
        \mu(x)_{b,c} = \frac1{HW} \sum_{h=1}^H{\sum_{w=1}^W{x_{b,c,h,w}}},
        \\
        \sigma^2(x)_{b,c} = \frac1{HW} \sum_{h=1}^H{\sum_{w=1}^W{{\left( x_{b,c,h,w} - \mu(x)_{b,c} \right)}^2}} + \eta,
    \end{gathered}
\end{equation}
where \begin{small}$\mu(x) = [\mu(x)_{b,c}]_{\substack{1 \le b \le B \\ 1 \le c \le C}}$\end{small}, \begin{small}$\sigma(x) = [\sigma(x)_{b,c}]_{\substack{1 \le b \le B \\ 1 \le c \le C}}$\end{small}, and $\eta > 0$ is a small value for numerical stability. We note that the $\AdaIN$ function in Eq.~\eqref{eqn:adain} first normalizes the style of $x$ by calculating $(x - \mu(x)) / \sigma(x)$, and then scales the result by $\sigma_t$ and shifts it by $\mu_t$ to move it to the desired target style.

\subsubsection{Higher-order statistics}
After computing the mean $\mu(x)$ and standard deviation $\sigma(x)$ for a given batch $x$ as its first-order style statistics, we can further extract second-order style statistics $\Sigma_\mu^2(x), \Sigma_\sigma^2(x) \in \mathbb{R}^C$ as the variance of each of them:
\vspace{-2mm}
\begin{equation} \label{eqn:Sigma_mu_Sigma_sig}
    \begin{gathered}
        \Sigma_\mu^2(x)_c = \frac1B \sum_{b=1}^B{{\left( \mu(x)_{b,c} - \mathbb{E}_b{[\mu(x)_{b,c}]} \right)}^2} + \eta,
        \\
        \Sigma_\sigma^2(x)_c = \frac1B \sum_{b=1}^B{{\left( \sigma(x)_{b,c} - \mathbb{E}_b{[\sigma(x)_{b,c}]} \right)}^2} + \eta,
    \end{gathered}
\end{equation}
where $\Sigma_\mu^2(x) = [\Sigma_\mu^2(x)_c]_{1 \le c \le C}$, $\Sigma_\sigma^2(x) = [\Sigma_\sigma^2(x)_c]_{1 \le c \le C}$.

\subsection{Problem Formulation} \label{ssec:problem}
We consider a decentralized device-to-device (D2D) network represented by a graph $\mathcal{G} = (\mathcal{M}, \mathcal{E})$, where $\mathcal{M} = \{ 1, ..., m \}$ is the set of devices/nodes, and $\mathcal{E}$ indicates the set of communication edges between the devices. We assume that the graph is connected and define the set of one-hop neighbors of a device $i \in \mathcal{M}$ as $\mathcal{N}_i$. In DFL~\cite{wang2024multi,chen2024taming}, device $i$ maintains an ML model $\theta_i$, and engages in (i) local model updates and (ii) inter-device communications to train a globally optimal model. Moreover, each device $i$ contains a local dataset $\mathcal{D}_i = \{ (x_i, y_i, d_i)_k \}_{k=1}^{|\mathcal{D}_i|}$ where the pair $(x_i, y_i, d_i)_k$ indicates the $k$th data point with input features $x_i$, label $y_i$, and unique to the DG setting, domain $d_i$.
For example, in the well-known PACS dataset \cite{li2017deeper}, each domain corresponds to a visual style, namely, art painting, cartoon, photo, or sketch.

In our DDG setting, the goal is to train a global domain-invariant model $\theta$ on the network $\mathcal{G}$ that generalizes well to unseen domains. In the PACS dataset example, a global model trained in the art painting, cartoon, and photo domains should generalize well to the sketch domain, despite the absence of sketch data and a centralized aggregator during training.

\subsection{Formal Modeling of Centralized DG} \label{ssec:cen_model}
In this section, we provide the first formal models of the $\MixStyle$ \cite{zhou2021domain} and $\DSU$ \cite{li2022uncertainty} algorithms for centralized DG. Note that the original papers and subsequent works do not present any such objective functions. This serves as a precursor to developing our \textsc{StyleDDG} algorithm in Sec.~\ref{ssec:styleddg}.

\subsubsection{$\MixStyle$ \cite{zhou2021domain}} In $\MixStyle$, the style space of the CNN layers is explored through convex combinations of the existing styles, thus exposing the model to styles beyond the existing ones in the training batch. For a given instance $x$ and target style statistics $\mu_t$ and $\sigma_t$, the $\MixStyle$ operation employs $\AdaIN$ from Eq.~\eqref{eqn:adain} as
\begin{equation} \label{eqn:mixstyle}
    \underset{\lambda}{\MixStyle}(x, \mu_t, \sigma_t) = \AdaIN(x, \beta_\lambda(x, \mu_t), \gamma_\lambda(x, \sigma_t)),
    \vspace{-0.1in}
\end{equation}
where the shifting and scaling parameters are defined as
\vspace{-2mm}
\begin{equation} \label{eqn:mixstyle_shift_scale}
    \begin{gathered}
        \beta_\lambda(x, \mu_t) = \lambda \mu(x) + (1-\lambda) \mu_t,
        \\
        \gamma_\lambda(x, \sigma_t) = \lambda \sigma(x) + (1-\lambda) \sigma_t,
    \end{gathered}
    \vspace{-2mm}
\end{equation}
respectively, and $\lambda$ is the mixing coefficient. The original paper \cite{zhou2021domain} applies $\MixStyle$ to all convolutional blocks except the last one for ResNet ($L = 3$), and samples $\lambda$ for each layer using the Beta distribution $\mathbb{p}_\lambda = \Beta(0.1, 0.1)$. Additionally, for each layer $\ell=1,...,L$, they assign a probability $p_\ell = 0.5$ of applying $\MixStyle$ to layer $\ell$.

\textbf{Our formal objective model.} We can formally model $\MixStyle$'s objective function. For a mini-batch $x$ of size $B$, $\MixStyle$'s loss function with random shuffling is given as
\vspace{-2mm}
\begin{equation} \label{eqn:mixstyle_loss}
    F(\theta) = \hspace{-2mm} \sum_{\pi \in \mathcal{P}(L)}{\hspace{-1mm} {\left( \prod_{\ell \in \pi}{p_\ell} \right)}  \mathbb{E}_{\substack{(x,y) \sim \mathbb{p}_D \\ \{ I_\ell \in \mathfrak{S}_\mathcal{B} \}_{\ell \in \pi} \\ \{ \lambda_\ell \sim \mathbb{p}_\lambda \}_{\ell \in \pi}} \hspace{-1mm}}{\left[ \mathscr{L}{(\hat{h}_{\theta, \{ I_\ell, \lambda_\ell \}_{\ell \in \pi}}(x), y)} \right]}},
    \vspace{-1mm}
\end{equation}
where $\mathbb{p}_D$ denotes the probability distribution of the data in the source domains and $\mathfrak{S}_\mathcal{B}$ is the set of all possible permutations of the set $\mathcal{B} = \{ 1,..., B \}$. In other words, \cite{zhou2021domain} applies the $\MixStyle$ strategy to the output of the model's layer $\ell$ with a probability $p_\ell$, so that the overall model is modified as
\begin{equation} \label{eqn:h_hat}
    \hat{h}_{\theta, \{ I_\ell, \lambda_\ell \}_{\ell \in \pi}}(x) = (\hat{h}_{\theta_L, I_L, \lambda_\ell} \circ \cdots \circ \hat{h}_{\theta_1, I_1, \lambda_1})(x),
\end{equation}
in which $\hat{h}_{\theta_\ell, I_\ell, \lambda_\ell} = h_{\theta_\ell}$ if $\ell \notin \pi$. If $\ell \in \pi$, we define $x_m = [x_k]_{k \in I_l}^T$ as a permutated version of the mini-batch $x$ based on the shuffled indices $I_\ell$, and then the $\MixStyle$ update that occurs in that layer can be defined using Eq.~\eqref{eqn:mixstyle} as
\begin{equation} \label{eqn:h_hat_l}
    \hat{h}_{\theta_\ell, I_\ell, \lambda_\ell}(x) = \underset{\lambda_\ell}{\MixStyle}(h_{\theta_\ell}(x), \mu(h_{\theta_\ell}(x_m)), \sigma(h_{\theta_\ell}(x_m))).
    \vspace{-1mm}
\end{equation}
Note that the $\MixStyle$ layer is taking the model output for mini-batch $x$ and the style statistics $\mu(\cdot)$ and $\sigma(\cdot)$ of the permuted mini-batch $x_m$ from Eq.~\eqref{eqn:mu_sig} for feature augmentation.

\subsubsection{$\DSU$ \cite{li2022uncertainty}} $\DSU$ is another centralized DG algorithm where the style space of the CNN layers are explored by adding a Gaussian noise to the existing styles. For a given instance $x$ and target higher-order style statistics $\Sigma_{\mu,t}$ and $\Sigma_{\sigma,t}$, the $\DSU$ operation invokes $\AdaIN$ from Eq.~\eqref{eqn:adain} as
\vspace{-2mm}
\begin{equation} \label{eqn:dsu}
    \begin{aligned}
        \underset{{\epsilon_\mu, \epsilon_\sigma}}{\DSU}(x, \Sigma_{\mu,t}, \Sigma_{\sigma,t}) = \AdaIN(x, & \beta_{\epsilon_\mu}(\mu(x), \Sigma_{\mu,t}),
        \\
        & \gamma_{\epsilon_\sigma}(\sigma(x), \Sigma_{\sigma,t})),
    \end{aligned}
    \vspace{-2mm}
\end{equation}
where the shifting and scaling parameters are defined as
\vspace{-1mm}
\begin{equation} \label{eqn:DSU_shift_scale}
    \beta_{\epsilon_\mu}(\mu, \Sigma_{\mu,t}) = \mu + \epsilon_\mu \Sigma_{\mu,t}, \; \gamma_{\epsilon_\sigma}(\sigma, \Sigma_{\sigma,t}) = \sigma + \epsilon_\sigma \Sigma_{\sigma,t},
\end{equation}
respectively, and $\epsilon_\mu$ and $\epsilon_\sigma$ follow a Normal distribution. Specifically, \cite{li2022uncertainty} applies $\DSU$ to all convolutional blocks along with the first convolutional layer and the max-pooling layer of a ResNet model ($L=6$). The exploration coefficients $\epsilon_{\mu,\ell}, \epsilon_{\sigma,\ell}$ are sampled for each layer $\ell$ from distributions $\mathbb{p}_{\epsilon_\mu}, \mathbb{p}_{\epsilon_\sigma}$ set as $\mathbb{p}_{\epsilon_\mu} = \mathbb{p}_{\epsilon_\sigma} = \mathcal{N}(0,1)$, i.e., unit Gaussian distribution. As in $\MixStyle$, the probability of applying $\DSU$ to each layer $\ell = 1, ..., L$ is set to $p_\ell = 0.5$.

\textbf{Our formal objective model.} For a mini-batch $x$ of size $B$, the loss function of $\DSU$ can be formalized as 
\vspace{-2mm}
\begin{equation} \label{eqn:dsu_loss}
    F(\theta) = \hspace{-3mm} \sum_{\pi \in \mathcal{P}(L)}{\hspace{-2mm} \left( \prod_{\ell \in \pi}{p_\ell} \right) \mathbb{E}_{\hspace{-3mm} \substack{(x,y) \sim \mathbb{p}_D \\ \{ \epsilon_{\mu,\ell} \sim \mathbb{p}_{\epsilon_\mu} \}_{\ell \in \pi} \\ \{ \epsilon_{\sigma,\ell} \sim \mathbb{p}_{\epsilon_\sigma} \}_{\ell \in \pi}} \hspace{-3mm}}{\left[ \mathscr{L}{(\dot{h}_{\theta, \{ \epsilon_{\mu, \ell}, \epsilon_{\sigma, \ell} \}_{\ell \in \pi}}(x), y)} \right]}}.
    \vspace{-1mm}
\end{equation}
In essence, this method applies its $\DSU$ strategy to the output of the model's layer $\ell$ with probability $p_\ell$, so that the overall model is modified as
\begin{equation} \label{eqn:h_dot}
    \dot{h}_{\theta, \{ \epsilon_{\mu,\ell}, \epsilon_{\sigma,\ell} \}_{\ell \in \pi}}(x) = (\dot{h}_{\theta_L, \epsilon_{\mu,L}, \epsilon_{\sigma,L}} \circ \cdots \circ \dot{h}_{\theta_1, \epsilon_{\mu,1}, \epsilon_{\sigma,1}})(x),
\end{equation}
in which $\dot{h}_{\theta_\ell, \epsilon_{\mu,\ell}, \epsilon_{\sigma,\ell}} = h_\theta$ if $\ell \notin \pi$. If $\ell \in \pi$, then the $\DSU$ update that occurs in that layer can be defined as
\begin{equation} \label{eqn:h_dot_l}
    \dot{h}_{\theta_\ell, \epsilon_{\mu,\ell}, \epsilon_{\sigma,\ell}}(x) = \underset{\epsilon_{\mu,\ell}, \epsilon_{\sigma,\ell}}{\DSU} \big( h_{\theta_\ell}(x), \Sigma_\mu{\left( h_{\theta_\ell}(x) \right)}, \Sigma_\sigma{\left( h_{\theta_\ell}(x) \right)} \big),
\end{equation}
where the $\DSU$ layer takes the model outputs for the mini-batch $x$ and the higher-order style statistics $\Sigma_\mu(\cdot)$ and $\Sigma_\sigma(\cdot)$ of that batch as given in Eq.~\eqref{eqn:Sigma_mu_Sigma_sig} to perform feature augmentation.

The loss function formulations we derived in Eqs.~\eqref{eqn:mixstyle_loss} and~\eqref{eqn:dsu_loss} provide us with compact representations that will enable us to formally model our \textsc{StyleDDG} algorithm in Sec.~\ref{sec:decen_dg}. We observe that minimizing Eqs.~\eqref{eqn:mixstyle_loss} and~\eqref{eqn:dsu_loss} results in learning style-invariant model parameters for all different combinations of layers/blocks in the CNN, by taking a convex combination of the styles (for $\MixStyle$) and adding Gaussian noise to the styles of each layer (for $\DSU$).

\section{Decentralized Style-Based DG} \label{sec:decen_dg}
In Sec.~\ref{ssec:algorithm}, we present the steps of our \textsc{StyleDDG} decentralized DG methodology, summarized in Alg.~\ref{alg:styleddg}. Then, we proceed to formally model \textsc{StyleDDG} in Sec.~\ref{ssec:styleddg}.

\subsection{\textsc{StyleDDG} Algorithm} \label{ssec:algorithm}

\subsubsection{Gradient Descent and Consensus Mechanism}
In our decentralized network setup, each device $i \in \mathcal{M}$ maintains a local model $\theta_i^{(k)}$ at each iteration $k$ of the training process. A consensus mechanism is necessary to ensure $\lim_{k \to \infty}{\theta_1^{(k)}} = ... = \lim_{k \to \infty}{\theta_m^{(k)}}$. To this end, we represent the update rule for each device $i \in \mathcal{M}$ in each iteration $k$ as in~\cite{zehtabi2024decentralized,chen2024taming}:
\begin{equation} \label{eqn:update}
    \theta_i^{(k+1)} = \theta_i^{(k)} + \sum_{j \in \mathcal{N}_i}{p_{ij} \left( \theta_j^{(k)} - \theta_i^{(k)} \right)} - \alpha^{(k)} g_i^{(k)},
    \vspace{-0.1in}
\end{equation}
where $p_{ij}$ is a mixing coefficient used for the model parameters of the device pair $(i,j)$ with a D2D link. Here, we employ the commonly used Metropolis-Hastings weights $p_{ij} = \min\{ 1 / (1 + |\mathcal{N}_i|), 1 / (1 + |\mathcal{N}_j|)\}$ \cite{xiao2004fast}. Furthermore, $g_i^{(k)} = \nabla{F}_i(\theta_i^{(k)}, [ \psi_j^{(k)} ]_{j \in \mathcal{N}_i})$ is the style-based gradient of device $i \in \mathcal{M}$ at iteration $k$ obtained based on the local loss function, which will be elaborated in Sec.~\ref{sssec:styleexplore}. It has been well-established how the convergence speed of DFL in~\eqref{eqn:update} decreases with the spectral radius $\rho$ of the mixing matrix on the D2D graph $\mathcal{G}$~\cite{zehtabi2024decentralized,koloskova2020unified}; a main contribution of our work is expanding such results to style sharing in DG.

\subsubsection{Style Statistics} \label{sssec:stylestatistics}
Let $X \in \mathbb{R}^{m \times B}$ be the concatenation of all local mini-batches of data from the devices, i.e., $X = \begin{bmatrix} x_1 & ... & x_m \end{bmatrix}^T$. We denote the set of styles from all the devices obtained using model parameters $\theta$ and the combined batch $X$ for all devices as $\Psi_\theta(X) = \begin{bmatrix} \psi_\theta(x_1) & \cdots & \psi_\theta(x_m) \end{bmatrix}^T$. The style tensor that each device $i \in \mathcal{M}$ calculates and shares with its neighbors is defined as $\psi_\theta(x_i) = \begin{bmatrix} \psi_{\theta,1}(x_i) & \cdots & \psi_{\theta,L}(x_i) \end{bmatrix}^T$, which is obtained for different CNN layers/blocks. Each individual style in this tensor consists of four style statistics at each layer/block \begin{small}$\psi_{\theta,\ell}(x_i) = \begin{bmatrix} \bar{\mu}_{\theta,\ell}(x_i) & \bar{\sigma}_{\theta,\ell}(x_i) & \Sigma_{\mu,\theta,\ell}^2(x_i) & \Sigma_{\sigma,\theta,\ell}^2(x_i) \end{bmatrix}^T$\end{small}:
\begin{equation} \label{eqn:style_statistics}
    \vspace{-0.1in}
    \begin{gathered}
        \bar{\mu}_{\theta,\ell}(x_i)_c = \frac1B \sum_{b=1}^B{\mu(h_{\theta_\ell}(x_i))_{b,c}},
        \\
        \bar{\sigma}_{\theta,\ell}(x_i)_c = \frac1B \sum_{b=1}^B{\sigma(h_{\theta_\ell}(x_i))_{b,c}},
        \\
        \Sigma_{\mu,\theta,\ell}^2(x_i)_c = \Sigma_\mu^2(h_{\theta_\ell}(x_i))_c, \;
        \Sigma_{\sigma,\theta,\ell}^2(x_i)_c = \Sigma_\sigma^2(h_{\theta_\ell}(x_i))_c,
    \end{gathered} \hspace{-1cm}
\end{equation}
where functions $\mu(\cdot)_{b,c}$ and $\sigma(\cdot)_{b,c}$ are given in Eq.~\eqref{eqn:mu_sig}, and functions $\Sigma_\mu^2(\cdot)_c$ and $\Sigma_\sigma^2(\cdot)_c$ are given in Eq.~\eqref{eqn:Sigma_mu_Sigma_sig}.

Therefore, the dimensions of the style vectors will be $\psi_{\theta_\ell}(x_i) \in \mathbb{R}^{4 \times C}$ for each layer, $\psi_\theta(x_i) \in \mathbb{R}^{L \times 4 \times C}$ for each device, and $\Psi_\theta(X) \in \mathbb{R}^{m \times L \times 4 \times C}$ for the whole network. Fig.~\ref{fig:stylestatistics} illustrates these style vectors and statistics.

\begin{remark}(Communication overhead)
    The added communication overhead for sharing style statistics $\psi_\theta(x_i) \in \mathbb{R}^{L \times 4 \times C}$ is negligible relative to the model parameters $\theta \in \mathbb{R}^n$. Take ResNet18 as an example, with around $n = 11.69$ million parameters. The length of the full style statistics vector in \textsc{StyleDDG} is just $1792$
    (four equidimensional style statistics in each layer, calculated in the outputs of the first three convolutional blocks with $64$, $128$ and $256$ channels, respectively), giving added communication overhead of just $0.015\%$.
\end{remark}

\subsubsection{Style Exploration Strategy} \label{sssec:styleexplore}

In \textsc{StyleDDG}, the goal of each device is to communicate style information with its neighbors so that they can collectively train domain-invariant models. Towards this, we apply three functions to the outputs of particular layers of a CNN, namely $\StyleShift$, $\StyleExplore$ and $\MixStyle$. In the following subsections, we explain the details of each of them.

\textbf{StyleShift. }
Each client first shifts the styles of $B_s \in \mathbb{Z}, B_s \in [0, B]$ of a mini-batch at a particular layer to the styles received from a neighbor. Illustrated in Fig.~\ref{fig:sysdiag}-(b), we call this the $\StyleShift$ layer. In a generic definition, $\StyleShift$ takes the batch \begin{small}$x = \begin{bmatrix} x_s & x_s^c \end{bmatrix}$\end{small} which has already been separated to two parts $x_s$ and $x_s^c$, and uses the style vector $\psi$ to shift a portion $B_s$ of the styles as
\begin{equation} \label{eqn:styleshift}
    \begin{gathered}
        \underset{\psi, \epsilon_\mu, \epsilon_\sigma}{\StyleShift}(x_s, x_s^c) = \begin{bmatrix} \tilde{x}_s & x_s^c \end{bmatrix}^T,
        \\
        \tilde{x}_s = \AdaIN(x_s, \beta_{\epsilon_{\mu,\ell}}(\mathbf{1}_{B_s} \bar{\mu}_{j_\ell}^T, \Sigma_{\mu,j_\ell}), \gamma_{\epsilon_{\sigma,\ell}}(\mathbf{1}_{B_s} \bar{\sigma}_{j_\ell}^T, \Sigma_{\sigma,j_\ell})),
    \end{gathered}
    \hspace{-8mm}
\end{equation}
$\mathbf{1}_{B_s}$ is an all-ones vector of size $B_s$, and $B$ is the batch size.
Recall that the function $\AdaIN$ was defined in Eq.~\eqref{eqn:adain}. The shifting and scaling hyperparameters are the same ones used for $\DSU$ as defined in Eq.~\eqref{eqn:DSU_shift_scale}.

\textbf{StyleExplore.} After shifting the styles of $B_s$ points, we perform another step called \textit{style exploration} to further expand the style space that each device is exposed to during training. To this end, we first calculate the intra-device average of style statistics, i.e., $\mathbb{E}_b[\mu(x)]$ and $\mathbb{E}_b[\sigma(x)]$. Then, we use those average points to extrapolate $B_e \in \mathbb{Z}, B_e \in [0, B]$ of the outputs of $\StyleShift$ outward of the style space, as illustrated in Fig.~\ref{fig:sysdiag}-(b). To mathematically model the points that are extrapolated, we let the function $\StyleExplore$ take a binary vector $I_e$ with $1^T I_e = B_e$ as input and shift those points in $x$ which correspond to $1$s in $I_e$.
In other words, for any given vector $x$, we choose $B_e$ of it $x_e =[x_k]_{k \in \{1,...,B\} ; (I_e)_k = 1}^T$ to apply style extrapolation on and keep the rest $x_e^c =[x_k]_{k \in \{1,...,B\} ; (I_e)_k = 1}^T$ the same. We finally concatenate them to get
\begin{equation} \label{eqn:explore}
    \begin{gathered}
        \mu_e = \begin{bmatrix}
            \mu(x_e) + \alpha (\mu(x_e) - \mathbb{E}_b[\mu(x)] & & \mu(x_e^c)
        \end{bmatrix}^T,
        \\
        \sigma_e = \begin{bmatrix}
            \sigma(x_e) + \alpha (\sigma(x_e) - \mathbb{E}_b[\sigma(x)]) & & \sigma(x_e^c)
        \end{bmatrix}^T,
    \end{gathered}
\end{equation}
where $\alpha$ is an exploration hyperparameter set to $\alpha=3$ as in \cite{park2024stablefdg}. Unlike \cite{park2024stablefdg}, we do not perform any oversampling in this layer to avoid doubling the computational overhead of our methodology. This is a very critical modification to make, since other well-known DG methods such as \cite{zhou2021domain, li2022uncertainty, lifedbn} all operate with the rationale that style exploration layers should be plug-and-play components of existing models.

\textbf{MixStyle.} As a final step, we use another set of indices $I_m$ (a $0$-$1$ vector similar to $I_e$) to obtain a permutation of $\mu_e$ and $\sigma_e$ and use them to do $\MixStyle$, with function $\MixStyle$ defined in Eq.~\eqref{eqn:mixstyle}. This step of our method is also illustrated in Fig.~\ref{fig:sysdiag}-(c). The final function can be written as
\begin{equation} \label{eqn:styleexplore}
    \begin{aligned}
        \underset{\lambda, I_e, I_m}{\StyleExplore}(x) \hspace{-0.5mm} = \hspace{-0.5mm} \underset{\lambda}{\MixStyle} \hspace{-0.5mm} \left( x, \begin{bmatrix} \mu_e \end{bmatrix}_{k \in I_m}^T \hspace{-1.5mm}, \begin{bmatrix} \sigma_e \end{bmatrix}_{k \in I_m}^T \right).
    \end{aligned}
\end{equation}

\begin{algorithm}[t]
    \small
    \caption{Decentralized Domain Generalization with Style Sharing (\textsc{StyleDDG})}
    \label{alg:styleddg}
    \DontPrintSemicolon
    \KwIn{$K$, $\mathcal{G} = (\mathcal{M}, \mathcal{E})$, ${\lbrace \alpha^{(k)} \rbrace}_{0 \le k \le K}$}
    \KwOut{${\lbrace \mathbf{\theta}_i^{(K+1)} \rbrace}_{i \in \mathcal{M}}$}
   
    $k \gets 0$, Initialize $\theta^{(0)}$, ${\lbrace \mathbf{\theta}_i^{(0)} \gets \theta^{(0)} \rbrace}_{i \in \mathcal{M}}$;
   
    \While{$k \le K$}{
        \ForAll{$i \in \mathcal{M}$}{
            sample mini-batch $(x_i^{(k)}, y_i^{(k)}) \in \mathcal{D}_i$; \label{alg:styleddg:minibatch}
            
            calculate $\psi_i^{(k)} = \psi_{\theta_i^{(k)}}(x_i^{(k)})$ using Eq.~\eqref{eqn:style_statistics};
            share $\theta_i^{(k)}$ and $\psi_i^{(k)}$ with neighbors $\mathcal{N}_i$;
        }
    
        \ForAll{$i \in \mathcal{M}$}{
    
            use the same mini-batch $(x_i^{(k)}, y_i^{(k)})$ from Line \ref{alg:styleddg:minibatch};
            
            $g_i^{(k)} \gets \nabla{F}_i(\theta_i^{(k)}, [ \psi_j^{(k)} ]_{j \in \mathcal{N}_i})$ (with respect to functions explained in Sec.~\ref{sssec:styleexplore});
            
            $\text{aggr}_i^{(k)} \gets 0$;
    
            \ForAll{$j \in \mathcal{E}_i$}{
                $p_{ij} \gets 1 / (1 + \max{\lbrace |\mathcal{N}_i|, |\mathcal{N}_j| \rbrace})$;
                
                $\text{aggr}_i^{(k)} \gets \text{aggr}_i^{(k)} + p_{ij} \left( \mathbf{\theta}_j^{(k)} - \mathbf{\theta}_i^{(k)} \right)$;
            }
       }
    
       \For{$i \in \mathcal{M}$}{
            $\mathbf{\theta}_i^{(k+1)} \gets \mathbf{\theta}_i^{(k)} + \text{aggr}_i^{(k)} - \alpha^{(k)} g_i^{(k)}$;
        }
        $k \gets k + 1$;
    }
\end{algorithm}

\textbf{Novelty Summary.} In a nutshell, the novelty of our algorithm is two-fold. First,  \textsc{StyleDDG} is, to our knowledge, the first style-based algorithm for achieving DG over fully decentralized D2D networks. The consensus-based decentralized update rule in Eq.~\eqref{eqn:update} is used with modified gradients $\nabla{F}_i(\theta_i^{(k)}, [ \psi_j^{(k)} ]_{j \in \mathcal{N}_i})$ to achieve DG. Second, our style-based approach utilizes the sharing of style information only between the one-hop neighbors of a device $i \in \mathcal{M}$, and is still able to train style-invariant ML models, as we will see next.

\subsection{Formal Modeling of \textsc{StyleDDG}} \label{ssec:styleddg}
Since our goal is to analyze the performance of all devices in the decentralized setup, we consider both a global loss and a device-side local loss for \textsc{StyleDDG}. This leads to one of the first concrete formulations among all style-based DG papers (centralized and star-topology federated algorithms included), with compact loss function representations.

\subsubsection{Global Loss}
Given the set of devices $\mathcal{M} = \{ 1, ..., m \}$, the global loss function of \textsc{StyleDDG} can be formulated as
\vspace{-2mm}
\begin{equation} \label{eqn:styleddg_global_loss}
    F(\theta) = \frac1m \sum_{i=1}^m{\mathbb{E}_{\{ (x_j, y_j) \sim \mathbb{p}_{D_j} \}_{j \in \mathcal{M}}}{\left[ F_i(\theta, \Psi_{\theta, i}(X)) \right]}},
    \vspace{-2mm}
\end{equation}
where $F_i(\theta, \Psi_{\theta,i}(X))$ is the local loss function of each device $i \in \mathcal{M}$ evaluated with the model $\theta$ and the combined batch $X$ over devices, in which $\Psi_{\theta,i}(X) \subset \Psi_\theta(X)$ is the set of styles belonging to the neighbors of device $i$. Note that in Eq.~\eqref{eqn:styleddg_global_loss}, if instead of $\Psi_{\theta,i}(X)$ we provide the full style information of $\Psi_{\theta}(X)$ to each individual device, we recover the loss function for DG in star-topology FL, as considered in StableFDG \cite{park2024stablefdg}.

\subsubsection{Local Loss}
We define the device-side local loss as
\vspace{-1mm}
\begin{equation} \label{eqn:styleddg_local_loss}
    F_i(\theta, \Psi) \hspace{-1mm} = \hspace{-4mm} \sum_{\pi \in \mathcal{P}(L)}{\hspace{-2mm} \left( \prod_{\ell \in \pi}{p_\ell} \hspace{-1mm} \right) \hspace{-1mm} \mathbb{E}_{\hspace{-3mm} \substack{(x,y) \sim \mathbb{p}_{D_i} \\ \{ j_\ell \in \mathcal{M} \setminus \{ i \} \}_{\ell \in \pi} \\ \{ \mathcal{R}_\ell \sim \mathbb{p}_{\mathcal{R}} \}_{\ell \in \pi}} \hspace{-4mm}}{\left[ \mathscr{L}{(\tilde{h}_{\theta, \{ \psi_{j_\ell}, \mathcal{R}_\ell \}_{l \in \pi}}(x), y)} \right]}} \hspace{-0.5mm},
    \vspace{-2mm}
\end{equation}
in which the expectation is taken over the set of random variables $\mathcal{R}_{\ell} = \left\{ \epsilon_{\mu, \ell}, \epsilon_{\sigma, \ell}, \lambda_\ell, I_{s,\ell}, I_{e,\ell}, I_{m,\ell} \right\}$, and the entries are sampled from the corresponding set of distributions $\mathbb{p}_{\mathcal{R}} = \{ \mathbb{p}_{\epsilon_\mu}, \mathbb{p}_{\epsilon_\sigma}, \mathbb{p}_\lambda, \mathfrak{S}_{\mathcal{I}_s}, \mathfrak{S}_{\mathcal{I}_e}, \mathfrak{S}_\mathcal{B} \}$. $\mathbb{p}_{D_i}$ denotes the probability distribution of source data in device $i \in \mathcal{M}$, and $\mathfrak{S}_{\mathcal{I}_s}$ and $\mathfrak{S}_{\mathcal{I}_e}$ are supersets of all unique permutations of the sets $\mathcal{I}_s$ and $\mathcal{I}_e$, respectively. In turn, $\mathcal{I}_s$ and $\mathcal{I}_e$ are binary vectors of size $B$, with the number of $1$'s equal to $B_s$ and $B_e$, respectively. For a given layer $\ell$, each device $i$ uses the received styles of a neighboring device $j$ in the training process, i.e., \begin{small}$\psi_{j_\ell} = \begin{bmatrix} \bar{\mu}_{j_\ell} & \bar{\sigma}_{j_\ell} & \Sigma_{\mu, j_\ell}^2 & \Sigma_{\sigma, j_\ell}^2 \end{bmatrix}^T$\end{small} as the $j$th element of $\Psi$ in Eq.~\eqref{eqn:styleddg_local_loss}. In essence, \textsc{StyleDDG} applies a \textit{style exploration} strategy to the output of the model's layer $\ell$ with probability $p_\ell$, so that the overall model is modified as
\begin{equation} \label{eqn:h_tilde}
    \tilde{h}_{\theta, \left\{ \psi_{j_\ell}, \mathcal{R}_\ell \right\}_{\ell \in \pi}} \hspace{-4mm} (x) = \left( \tilde{h}_{\theta_L, \psi_{j_L}, \mathcal{R}_L} \circ \cdots \circ \tilde{h}_{\theta_1, \psi_{j_1}, \mathcal{R}_1} \right)(x),
\end{equation}
in which $\tilde{h}_{\theta_\ell, \psi_{j_\ell}, \mathcal{R}_\ell} = h_{\theta_\ell}$ if $\ell \notin \pi$. If $\ell \in \pi$, we define $x_s =[x_k]_{k \in \{1,...,B\} ; (I_{s,\ell})_k = 1}^T$ as a subset of the mini-batch $x$ based on the indices $I_{s,\ell}$ that we will shift to the styles of a neighboring device and $x_s^c = [x_k]_{k \in \{1,...,B\} ; (I_{s,\ell})_k = 0}^T$ as the ones which will not be shifted. Then the $\StyleExplore$ update that occurs in that layer can be expressed as
\vspace{-2mm}
\begin{equation} \label{eqn:h_tilde_l}
    \tilde{h}_{\theta_\ell, \psi_{j_\ell}, \mathcal{R}_\ell}(x) \hspace{-1mm} = \hspace{-1mm} \underset{\lambda_\ell, I_{m,\ell}, I_{e,\ell}}{\StyleExplore} \hspace{-1mm} \left( \hspace{-1mm} \underset{\psi_{j_\ell}, \epsilon_{\mu,\ell}, \epsilon_{\sigma,\ell}}{\StyleShift}(h_{\theta_\ell}(x_s), h_{\theta_\ell}(x_s^c))\hspace{-1mm} \right) \hspace{-1.5mm},
\end{equation}
where we first apply the $\StyleShift$ function on a given mini-batch $x$ to randomly shift $B_s$ of the points to new styles received from a neighbor, and then the $\StyleExplore$ layer takes the output of the $\StyleShift$ function to randomly extrapolate $B_e$ of those styles outwards for style exploration.

\section{Convergence Analysis} \label{sec:convergence}

In this section, we characterize the convergence behavior of \textsc{StyleDDG}. Papers such as \cite{koloskova2020unified, zehtabi2024decentralized, xin2021improved} prove that in decentralized optimization of general non-convex models, convergence to stationary points can be guaranteed if the local loss functions are smooth, i.e., their gradients are Lipschitz continuous. To this end, we will discuss the conditions under which the smoothness of local loss functions in \textsc{StyleDDG} given in Eq.~\eqref{eqn:styleddg_local_loss} can be guaranteed as well. In other words, our goal is to prove that the upper bound $\| \nabla{F}_i(\theta, \Psi) - \nabla{F}_i(\theta', \Psi') \| \le \beta_i \| \theta - \theta' \|$ can be established for each device in \textsc{StyleDDG}, where $\beta_i \in \mathbb{R}^+$ is the smoothness coefficient.

Before we proceed, note that our main local loss function given in Eq.~\eqref{eqn:styleddg_local_loss} essentially modifies the loss functions in a way that DG is achieved among the devices. We emphasize this to clarify what we mean by convergence in this context: whether the devices converge to the new optimal model with respect to the \textit{modified} local loss functions.

\subsection{Assumptions}
\begin{assumption} (Multivariate smoothness) \label{assump:smooth}
    The local loss functions in Eq.~\eqref{eqn:styleddg_local_loss} are smooth in their arguments, i.e.,
    \begin{equation} \label{eqn:smooth}
        \| \nabla{F}_i(\theta, \Psi) - \nabla{F}_i(\theta', \Psi') \| \le \beta_{\theta,i} \| \theta - \theta' \| + \beta_{\Psi,i} \| \Psi - \Psi' \|,
    \end{equation}
    in which $\beta_{\theta,i}, \beta_{\Psi,i} \in \mathbb{R}^+$ are positive scalars, $\theta, \theta' \in \mathbb{R}^n$ are any two model parameters and $\Psi, \Psi' \in \mathbb{R}^{|\mathcal{N}_i| \times L \times 4 \times C}$ are any two style vectors, for all $i \in \mathcal{M}$.
\end{assumption}

The Assumption~\ref{assump:smooth} is necessary to ensure the convergence of gradient-based methods, and it ensures that the local gradients do not change arbitrarily fast \cite{bottou2018optimization}. The conventional smoothness assumption made in \cite{bottou2018optimization, koloskova2020unified} is of the form $\| \nabla{F}_i(\theta) - \nabla{F}_i(\theta') \| \le \beta_{\theta,i} \| \theta - \theta' \|$. These works do not modify their CNN models with any style-based statistics, and the local loss functions are parameterized by $\theta$ alone. Assumption~\ref{assump:smooth} extends this to incorporate smoothness in terms of style vectors as well, i.e., the local loss functions do not change arbitrarily quickly with the style statistics.

\begin{remark} (Multivariate smoothness justification)
    If we concatenate $\theta$ and $\Psi$ to form the vector $\phi = [\theta^T, \Psi^T]^T$, then the left-hand side of Eq.~\eqref{eqn:smooth} can also be bounded as
    \begin{equation} \label{eqn:univariate_smooth}
        \| \nabla{F}_i(\phi) - \nabla{F}_i(\phi') \| \le \beta_{\phi,i} \| \phi - \phi' \|,
    \end{equation}
    which is the standard smoothness assumption for a univariate function with $\beta_{\phi,i} \in \mathbb{R}^+$. Note that Eq.~\eqref{eqn:univariate_smooth} is stricter than the multivariate smoothness assumption in Eq.~\eqref{eqn:smooth} since we can obtain the former from the latter using the triangle inequality.
\end{remark}

\begin{assumption} \label{assump:h}
    For each device $i \in \mathcal{M}$ and layer $\ell \in \{ 1,..., L \}$ of the CNN model $\theta$, the following holds for the outputs $h_{\theta_\ell}(x_i)$ given an input $x_i$ for device $i$:
    \begin{enumerate}[label=(\alph*),topsep=0pt]
        \item (Liptschitz continuity of layer outputs) We have
        $\| h_{\theta_\ell}(x_i)_{b,c,h,w} - h_{\theta_\ell'}(x_i)_{b,c,h,w} \| \le D_{i,\ell} \| \theta - \theta' \|,$
        where $\theta, \theta' \in \mathbb{R}^n$ are any two model parameter vectors and $D_{i, \ell} \in \mathbb{R}^+$ is a positive scalar. We also define $D_i = \max_\ell\{ D_{i,\ell} \}$ and $D = \max_i\{ D_i \}$. \label{assump:h:lipschitz}

        \item (Bounded layer output) For all $\theta \in \mathbb{R}^n$, there exists $U_{i,\ell} \in \mathbb{R}^+$ such that
        $\| h_{\theta_\ell}(x_i)_{b,c,h,w} \| \le U_{i,\ell} < \infty$.
        We also define $U_i = \max_\ell\{ U_{i,\ell} \}$ and $U = \max_i\{ U_i \}$. \label{assump:h:bounded}
    \end{enumerate}
\end{assumption}
Assumption~\ref{assump:h}-\ref{assump:h:lipschitz} indicates that changes in the model parameters $\theta$ do not lead to arbitrary large changes in the outputs of any layer. This is a safe assumption as the operations in CNNs (convolutional blocks, pooling layers, activation functions) behave this way. Assumption~\ref{assump:h}-\ref{assump:h:bounded} depends on the particular activation functions; for example, it holds for sigmoid, tanh and other bounded activations. With ReLU and other unbounded activations, for the purpose of our analysis, we only need this assumption to hold during the training phase.



\subsection{Main Results}
We begin with Propositions~\ref{proposition:lipschitz} and \ref{proposition:smooth}, key results which provide guarantees for the smoothness of local loss functions in \textsc{StyleDDG}. Subsequently, in Theorem~\ref{thm:convergence}, we use the propositions to show the convergence of \textsc{StyleDDG} for general non-convex loss functions.
In doing so, we ultimately prove that the second term in our smoothness assumption given in Eq.~\eqref{eqn:smooth}, i.e., $\| \Psi - \Psi' \|$, can also be bounded by $\| \theta - \theta' \|$.

\begin{proposition} (Lipschitz continuity of style statistics) \label{proposition:lipschitz}
    Let Assumption~\ref{assump:h} hold. Then the style statistics from Eq.~\eqref{eqn:style_statistics} are Lipschitz continuous at each layer/block $\ell = \{ 1, ..., L \}$ of CNN $\theta$ for all devices $i \in \mathcal{M}$, and for all channels $c \in \{1, ..., C\}$. Specifically:
    \begin{enumerate}[label=(\alph*)]
        \item $\| \bar{\mu}_{\theta, \ell}(x_i)_c - \bar{\mu}_{\theta', \ell}(x_i)_c \| \le D_{i,\ell} \| \theta - \theta' \|$. \label{proposition:lipschitz:mu}

        \item $\| \bar{\sigma}_{\theta, \ell}(x_i)_c - \bar{\sigma}_{\theta', \ell}(x_i)_c \| \le \frac{4 U_i D_{i,\ell}}{\sqrt{\eta}} \| \theta - \theta' \|$. \label{proposition:lipschitz:sigma}

        \item $\| \Sigma_{\mu, \theta, \ell}^2(x_i)_c - \Sigma_{\mu, \theta', \ell}^2(x_i)_c \| \le 4 U_i D_{i,\ell} \| \theta - \theta' \|$. \label{proposition:lipschitz:Sigma_mu}

        \item $\| \Sigma_{\sigma, \theta, \ell}^2(x_i)_c \!-\!\Sigma_{\sigma, \theta', \ell}^2(x_i)_c \| \!\le\! 4 U_i D_{i,\ell} \big(1\!+\!\frac{2\sqrt{2} U_i}{\sqrt{\eta}} \big) \| \theta \!-\! \theta' \|$. \label{proposition:lipschitz:Sigma_sigma}
    \end{enumerate}

    \textit{Proof.} See our technical report \cite{zehtabi2025decentralized}.
    
\end{proposition}
Proposition~\ref{proposition:lipschitz} shows that if the style statistics given in Eq.~\eqref{eqn:style_statistics} are calculated via the model $\theta$ on any data batch $X$, the resulting statistics will be Lipschitz with respect to $\theta$. With this result in hand, we can now prove the smoothness of local and global loss functions given in Eqs.~\eqref{eqn:styleddg_local_loss} and \eqref{eqn:styleddg_global_loss}.

\begin{table}[t]
    \setlength{\tabcolsep}{1.5pt}
    \centering
    \begin{sc}
        \begin{tabular}{c|c|c|c|c|r}
            \toprule
            \multirow{2}{*}{Method} & \multicolumn{5}{c}{Accuracy for Target Domain} \\
            & Art & Cartoon & Photo & Sketch & Avg \\
            \midrule
            FedBN & 53.6 $\pm$ 0.9 & 48.9 $\pm$ 1.6 & 74.0 $\pm$ 1.9 & 53.4 $\pm$ 5.5 & 57.5 \\
            FedBN + $\MixStyle$ & 59.7 $\pm$ 1.5 & 55.5 $\pm$ 1.7 & 78.2 $\pm$ 1.0 & 55.7 $\pm$ 3.0 & 62.3 \\
            FedBN + $\DSU$ & 59.5 $\pm$ 1.0 & 55.9 $\pm$ 2.5 & 79.0 $\pm$ 1.4 & 59.7 $\pm$ 3.9 & 63.5 \\
            \midrule
            DSGD & 67.7 $\pm$ 2.3 & 67.8 $\pm$ 1.1 & 91.1 $\pm$ 0.6 & 53.0 $\pm$ 9.9 & 69.9 \\
            DSGD + $\MixStyle$ & 74.7 $\pm$ 1.0 & 71.1 $\pm$ 2.1 & 91.1 $\pm$ 0.5 & 58.9 $\pm$ 4.1 & 73.9 \\
            DSGD + $\DSU$ & 76.6 $\pm$ 0.7 & 71.8 $\pm$ 1.0 & 91.7 $\pm$ 0.2 & 63.6 $\pm$ 3.0 & 75.9 \\
            \midrule
            \textsc{StyleDDG} & \textbf{77.7 $\pm$ 0.4} & \textbf{73.0 $\pm$ 0.4} & \textbf{94.2 $\pm$ 0.3} & \textbf{66.2 $\pm$ 0.6} & \textbf{77.8} \\
            \bottomrule
        \end{tabular}
    \end{sc}
    \caption{Results on the PACS dataset, for a network of $m=3$ devices all connected to each other in a full mesh topology. \vspace{-0.25in}}\label{tab:results1}
\end{table}

\begin{proposition} (Univariate smoothness) \label{proposition:smooth}
    Let Assumptions~\ref{assump:smooth} and \ref{assump:h} hold. Then for any $\theta, \theta' \in \mathbb{R}^n$, it holds that
        for any given combined batch $X$ across devices, the local loss function of each device $i \in \mathcal{M}$ is smooth, i.e.,
        \vspace{-2mm}
        $$\| \nabla{F}_i(\theta, \Psi_{\theta,i}(X)) - \nabla{F}_i(\theta', \Psi_{\theta',i}(X)) \| \le \beta_i \| \theta - \theta' \|,
        \vspace{-2mm}$$
        where $\beta_i = \beta_{\theta,i} + mL ( 1 + 4U ( \frac1{\sqrt{\eta}} + 2 + \frac{2\sqrt{2} U}{\sqrt{\eta}} ) ) D \beta_{\Psi,i}$,
        $\beta_{\theta,i}$ and $\beta_{\Psi,i}$ are provided in Assumption~\ref{assump:smooth}, $m$ is the total number of devices, $L$ is the number of layers that \textsc{StyleDDG} is applied to, $U$ and $D$ are the bounds of Assumption~\ref{assump:h}, and $\eta > 0$ is the small numerical stability constant. \label{proposition:smooth:local}


    \textit{Proof.} See our technical report \cite{zehtabi2025decentralized}.
    
\end{proposition}
In Proposition~\ref{proposition:smooth}, note that smaller $U = \max_{i,\ell}{U_{i,\ell}}$, smaller $D = \max_{i,\ell}{D_{i,\ell}}$, and larger $\eta$ each result in a smaller smoothness parameter $\beta$. This corresponds to a faster convergence during training.

\begin{theorem} \label{thm:convergence}
    Let Assumptions~\ref{assump:smooth} and \ref{assump:h} hold, and a constant learning rate $\alpha^{(k)} = \alpha^{(0)}/{\sqrt{K+1}}$ be employed where $K$ is the maximum number of iterations that \textsc{StyleDDG} will run for. Then, for a general non-convex loss function, we have
    \begin{equation} \label{eqn:convergence}
        \begin{aligned}
            & \frac{\sum_{k=0}^K{{\| \nabla{F}(\bar{\theta}^{(k)}) \|}^2}}{K+1} \le \frac{F(\bar{\theta}^{(0)}) - F^\star}{\mathcal{O}(\sqrt{K+1})}
            \\
            & \qquad \qquad \qquad + \frac{1+\rho}{(1 - \rho)^2} \frac{\beta^2}{\mathcal{O}{(K+1)}} + \frac{\beta/(2m)}{\mathcal{O}(\sqrt{K+1})},
        \end{aligned}
    \end{equation}
    where $\beta = (1/m) \sum_{i=1}^m{\beta_i}$, $F^\star = \min_{\theta \in \mathbb{R}^n}\{F(\theta)\}$, $m$ denotes the number of devices, $\bar{\theta}^{(k)} = (1/m) \sum_{i=1}^m{\theta_i^{(k)}}$ is the average model among all devices in iteration $k$, $\rho$ is the spectral radius of the connected network graph $\mathcal{G}$, and $\beta$ is the smoothness parameter given by Proposition~\ref{proposition:smooth}.

    \textit{Proof.} See our technical report \cite{zehtabi2025decentralized}.
    
\end{theorem}
Theorem~\ref{thm:convergence} shows that \textsc{StyleDDG} reaches a stationary point for general non-convex models at a rate of $\mathcal{O}(1 / \sqrt{K})$. To be specific, after running \textsc{StyleDDG} for $K$ iterations with a learning rate proportional to $1 / \sqrt{K+1}$, the running average ($\frac1{K+1} \sum_{k=0}^K{(\cdot)}$) of global gradient norm expectations ($\mathbb{E}{[\| \nabla{F}(\cdot) \|]}$), evaluated at the average model $\bar{\theta}^{(k)}$, is proportional to $1 / \sqrt{K+1}$. This bound becomes arbitrarily small as training progresses, i.e., as $K$ increases. Additionally, in Theorem~\ref{thm:convergence}, ML model-related parameters of Assumptions~\ref{assump:smooth} and \ref{assump:h}, i.e., multivariate smoothness parameters $\beta_{\theta, i}$ and $\beta_{\Psi, i}$, Lipschitz constant $D_{i,\ell}$ and the upper bound $U_{i,\ell}$, are all captured in the univariate smoothness parameter $\beta$ derived in Proposition~\ref{proposition:smooth}. We see in Eq.~\eqref{eqn:convergence} that a lower smoothness value $\beta$, as well as a better-connected D2D graph (i.e., smaller spectral radius $\rho$), results in a smaller upper bound in Eq.~\eqref{eqn:convergence}, accelerating the convergence speed accordingly.

\section{Numerical Experiments} \label{sec:experiments}

\begin{table}[t]
    \setlength{\tabcolsep}{1.5pt}
    \centering
    
    \begin{sc}
        \begin{tabular}{c|c|c|c|c|r}
            \toprule
            \multirow{2}{*}{Method} & \multicolumn{5}{c}{Accuracy for Target Domain} \\
            & Art & Cartoon & Photo & Sketch & Avg \\
            \midrule
            FedBN & 40.2 $\pm$ 0.9 & 35.5 $\pm$ 0.3 & 52.7 $\pm$ 1.4 & 34.0 $\pm$ 0.7 & 40.6 \\
            FedBN + $\MixStyle$ & 42.2 $\pm$ 1.0 & 39.1 $\pm$ 0.3 & 53.6 $\pm$ 1.4 & 38.6 $\pm$ 0.2 & 43.4 \\
            FedBN + $\DSU$ & 42.6 $\pm$ 0.8 & 40.8 $\pm$ 0.8 & 56.8 $\pm$ 2.4 & 39.3 $\pm$ 0.6 & 44.9 \\
            \midrule
            DSGD & 58.8 $\pm$ 1.1 & 52.5 $\pm$ 2.1 & 78.0 $\pm$ 2.0 & 36.7 $\pm$ 5.2 & 56.5 \\
            DSGD + $\MixStyle$ & \textbf{60.8 $\pm$ 1.4} & 55.6 $\pm$ 1.9 & 73.3 $\pm$ 0.9 & 39.9 $\pm$ 4.9 & 57.4 \\
            DSGD + $\DSU$ & 56.8 $\pm$ 1.2 & 54.1 $\pm$ 1.7 & 78.2 $\pm$ 3.4 & 35.5 $\pm$ 4.8 & 56.1 \\
            \midrule
            \textsc{StyleDDG} & 60.6 $\pm$ 1.6 & \textbf{56.3 $\pm$ 2.0} & \textbf{81.5 $\pm$ 4.0} & \textbf{42.3 $\pm$ 4.5} & \textbf{60.2} \\
            \bottomrule
        \end{tabular}
    \end{sc}
    \caption{Results on the PACS dataset, for a network of $m=9$ devices on a random geometric graph with radius $r = 0.8$. \vspace{-0.1in}}\label{tab:results2}
\end{table}

\subsection{Setup}
\textbf{Datasets and Model.} Our experiments are based on two widely-used datasets for benchmarking DG; (i) PACS \cite{li2017deeper} which consists of $4$ domains (art painting (A), cartoon (C), photo (P) and sketch (S)) and contains $7$ classes of images (dog, elephant, giraffe, guitar, horse, house, and person); and (ii) VLCS \cite{torralba2011unbiased} with $4$ domains (Caltech101, LabelMe, PASCAL VOC2007, and SUN09) and $5$ categories of images (bird, car, chair, dog, and person). For both datasets, we use $3$ source domains and hold out the remaining one as the target domain (for evaluating the test performance). By default, we conduct our experiments using the ResNet18 model \cite{he2016deep}.

\begin{table}[t]
    \setlength{\tabcolsep}{1.5pt}
    \centering
    \begin{sc}
        \begin{tabular}{c|c|c|c|c|r}
        \toprule
            \multirow{2}{*}{Method} & \multicolumn{5}{c}{Accuracy for Target Domain} \\
            & Caltech & Labelme & Pascal & Sun & Avg \\
            \midrule
            FedBN & 95.5 $\pm$ 0.8 & 56.6 $\pm$ 0.5 & 65.1 $\pm$ 0.7 & 66.1 $\pm$ 0.6 & 70.8 \\
            FedBN + $\MixStyle$ & 96.7 $\pm$ 0.5 & 57.2 $\pm$ 0.5 & 65.8 $\pm$ 0.4 & 66.9 $\pm$ 0.7 & 71.6 \\
            FedBN + $\DSU$ & 97.4 $\pm$ 0.2 & 56.5 $\pm$ 0.3 & 66.4 $\pm$ 0.3 & 67.1 $\pm$ 0.4 & 71.9 \\
            \midrule
            DSGD & 96.3 $\pm$ 0.8 & 56.0 $\pm$ 0.5 & 67.9 $\pm$ 0.6 & 68.5 $\pm$ 1.0 & 72.2 \\
            DSGD + $\MixStyle$ & 98.0 $\pm$ 0.6 & 56.9 $\pm$ 0.7 & 67.7 $\pm$ 1.8 & \textbf{70.1 $\pm$ 1.4} & 73.2 \\
            DSGD + $\DSU$ & 98.4 $\pm$ 0.3 & 56.1 $\pm$ 0.3 & 68.6 $\pm$ 0.6 & 69.9 $\pm$ 1.4 & 73.2 \\
            \midrule
            \textsc{StyleDDG} & \textbf{98.6 $\pm$ 0.8} & \textbf{58.1 $\pm$ 1.2} & \textbf{69.4 $\pm$ 0.1} & 69.4 $\pm$ 1.4 & \textbf{73.9} \\
            \bottomrule
        \end{tabular}
        \caption{Results on the VLCS dataset, for a network of $m=3$ devices all connected to each other. \vspace{-0.1in}}\label{tab:results3}
    \end{sc}
\end{table}
\begin{table}[t]
    \setlength{\tabcolsep}{1.5pt}
    \centering
    \begin{sc}
        \begin{tabular}{c|c|c|c|c|r}
            \toprule
            \multirow{2}{*}{Method} & \multicolumn{5}{c}{Accuracy for Target Domain} \\
            & Art & Cartoon & Photo & Sketch & Avg \\
            \midrule
            FedBN & 61.5 $\pm$ 0.6 & 51.8 $\pm$ 1.2 & 85.9 $\pm$ 1.1 & 59.5 $\pm$ 1.1 & 64.7 \\
            FedBN + $\MixStyle$ & 63.7 $\pm$ 0.8 & 60.1 $\pm$ 0.8 & 86.8 $\pm$ 0.5 & 65.0 $\pm$ 2.1 & 68.9 \\
            FedBN + $\DSU$ & 69.2 $\pm$ 1.7 & 59.1 $\pm$ 0.9 & 88.7 $\pm$ 1.1 & 69.1 $\pm$ 1.2 & 71.5 \\
            \midrule
            DSGD & 78.9 $\pm$ 0.4 & 75.2 $\pm$ 0.3 & 94.5 $\pm$ 0.3 & 65.8 $\pm$ 1.6 & 78.6
            \\
            DSGD + $\MixStyle$ & 81.8 $\pm$ 0.7 & \textbf{79.5 $\pm$ 0.6} & 94.7 $\pm$ 0.3 & 70.5 $\pm$ 0.9 & 81.6
            \\
            DSGD + $\DSU$ & 84.1 $\pm$ 0.5 & 77.2 $\pm$ 0.4 & 95.5 $\pm$ 0.0 & 74.4 $\pm$ 1.2 & 82.8
            \\
            \midrule
            \textsc{StyleDDG} & \textbf{87.5 $\pm$ 0.4} & 78.9 $\pm$ 0.3 & \textbf{97.4 $\pm$ 0.1} & \textbf{75.6 $\pm$ 1.0} & \textbf{84.9}
            \\
            \bottomrule
        \end{tabular}
        \caption{Results on the PACS dataset, for a network of $m=3$ devices all connected to each other using ResNet50. \vspace{-0.2in}}\label{tab:results4}
    \end{sc}
\end{table}

\textbf{Baselines.} Since our work is the first to propose a fully decentralized DG algorithm, we select some centralized and federated DG algorithms and implement a decentralized realization of them for comparison. Specifically, we compare \textsc{StyleDDG} against DSGD \cite{koloskova2020unified}, FedBN \cite{lifedbn}, $\MixStyle$ \cite{zhou2021domain} and $\DSU$ \cite{li2022uncertainty}. For the well-known style-based centralized DG methods, $\MixStyle$ and $\DSU$, we incorporate these algorithms into the local update process in each client.

\textbf{Setting.} Our experiments are done under two sets of networks: (i) $m = 3$ devices and (ii) $m = 9$ devices, connected together on a random geometric graph \cite{penrose2003random}. The choice of $3$-$9$ devices aligns with the experiment conventions found in FDG literature \cite{zhang2021federated, lifedbn, nguyen2022fedsr}, thus providing fair comparisons with the baselines. Our implementation of \textsc{StyleDDG} is based on the \textsc{dassl} library developed in \cite{zhou2022domain, zhou2021domainadaptive}. We adopt a learning rate $\alpha = 0.001$ with a cosine scheduler and use the batch size of $64$. We distribute the data among the edge devices in a domain-heterogeneous way, where each device gets samples only from a single domain. We measure the test accuracy of each algorithm after $50$ training epochs. We carried out the experiments using a cluster of four NVIDIA A100 GPUs with 40GB memory. We run the experiments five times in each setup and present the mean and standard deviation.

\subsection{Results and Discussion}
\textbf{PACS.} Table \ref{tab:results1} shows the results with the PACS dataset with $m=3$ clients. We make the following observations. First, existing style-based centralized DG methods, i.e., MixStyle and DSU, enhance the performance of both FedBN and DSGD by exposing models to a diversity of styles. However, applying these approaches does not allow the model to sufficiently explore diverse styles in decentralized setups, as each device has access to only a limited set of styles or domains. We observe that \textsc{StyleDDG} addresses this limitation through style sharing and exploration in decentralized settings, resulting in a significant improvement in generalization capability.

Table \ref{tab:results2} extends our results to $m = 9$ devices connected via a random geometric graph with radius $r = 0.8$. We see that the performance gain of \textsc{StyleDDG} over the baselines becomes more pronounced, highlighting its advantage in larger networks with less connectivity among devices.

\textbf{VLCS.} In Table \ref{tab:results3}, we evaluate performance on the VLCS dataset. Although the gains are smaller than those observed on PACS -- due to the relatively smaller style gap between domains in VLCS \cite{park2024stablefdg} -- \textsc{StyleDDG} still outperforms in all but one domain.

\begin{figure}[t]
    \vspace{-0.1in}
    \includegraphics[width=0.48\textwidth]{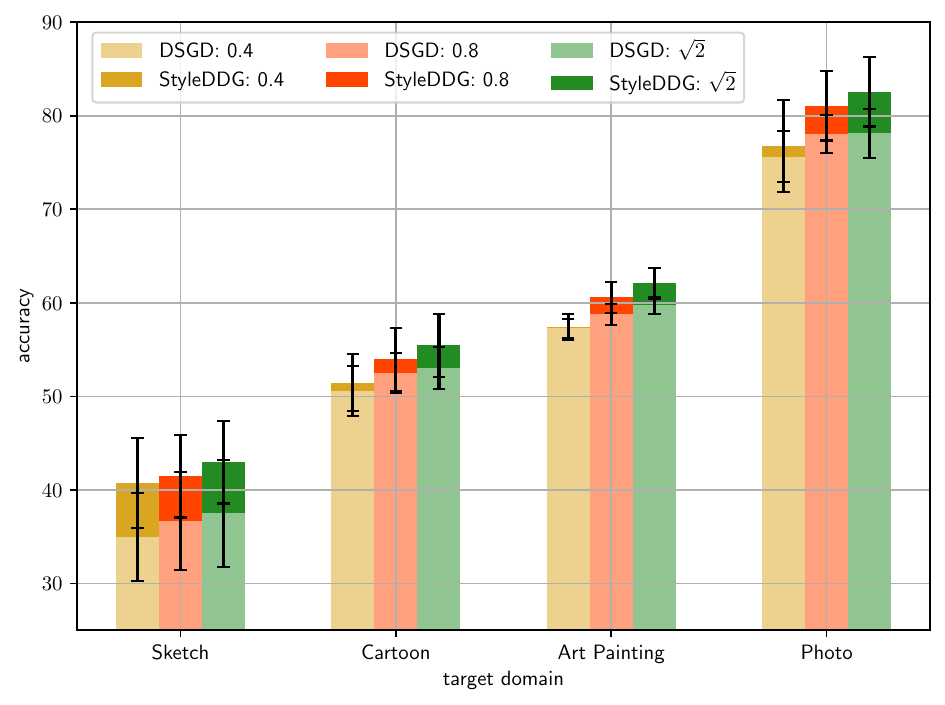}
    \vspace{-0.1in}
    \caption{Results on the PACS dataset for different target domains, for a network of $m=9$ clients over a random geometric graph with varying radius $r = 0.4, 0.8, \sqrt{2}$. \vspace{-0.25in}}
    \label{fig:graph_conn}
\end{figure}

\textbf{ResNet50.} In Table~\ref{tab:results4}, we explore the DG performance of our algorithm using a larger model, i.e., ResNet50 \cite{he2016deep}. Our results show that \textsc{StyleDDG} outperforms all baselines in all but one domain and is superior to them on average. Moreover, compared to the ResNet18 results given in Table~\ref{tab:results1}, we see a ${\small \sim} 7\%$ increase in accuracy for all DG algorithms, which shows that larger models lead to higher-quality style statistics for DG.

\textbf{Graph Connectivity.} In Fig.~\ref{fig:graph_conn}, we vary the radius of the random geometric graph in a network of $m=9$ devices and compare the performance of \textsc{StyleDDG} with DSDG for different levels of connectivity of the network. We observe that our methodology maintains its superiority across different connectivity levels. Furthermore, the advantage of our method increases with higher graph connectivity, showing that it can better utilize the underlying  network topology to achieve DG. In other words, the strength of style sharing in our method becomes more significant if each device is connected to more neighbors, as it can be exposed to a larger set of styles.

\section{Conclusion}
We proposed \textsc{StyleDDG}, to our knowledge the first decentralized federated domain generalization algorithm. \textsc{StyleDDG} enables edge devices in a decentralized network to train domain-invariant models. We achieved this by forming consensus-based model aggregation, style sharing, and developing a style exploration strategy to expose each device to the new style domains of their neighbors. Furthermore, we provided a novel analytical foundation of well-known domain generalization methods and extended it to our methodology, enabling us to conduct the first convergence analysis among all style-based domain generalization algorithms. Our experiments demonstrated the effectiveness of \textsc{StyleDDG} compared to other domain generalization baselines in terms of model accuracy on unseen target domains.

The authors have provided public access to their code and/or data at \href{https://zenodo.org/records/18204171}{https://zenodo.org/records/18204171}.

\vspace{-0.05in}
\section*{Acknowledgment}
This work was supported in part by the U.S. National Science Foundation (NSF) under grant No. CNS-2146171 and ECCS-2512911, the Office of Naval Research (ONR) under grant N00014-22-1-2305, the Defense Advanced Research Projects Agency (DARPA) under grant D22AP00168, and the Air Force Office of Scientific Research (AFOSR) under grant FA9550-24-1-0083.

\bibliographystyle{IEEEtran}
\bibliography{IEEEabrv, mybibfile}

\appendix

\section{Intermediary Lemma}

\begin{lemma} (Bounded $\mu$ and $\sigma$ style statistics) \label{lemma:bounds}
    Let Assumption~\ref{assump:h}-\ref{assump:h:bounded} hold. Then for all devices $i \in \mathcal{M}$ and all layers/blocks $\ell \in \{ 1, ..., L \}$ of the CNN $\theta$ it holds that
    \begin{enumerate}[label=(\alph*)]
        \item $\| \mu(h_{\theta_\ell}(x_i))_{b,c} \| \le U_{i,\ell}$, \label{lemma:bounds:mu}
        
        \item $\| \sigma(h_{\theta_\ell}(x_i))_{b,c} \| \le \sqrt{2} U_{i,\ell} + \sqrt{\eta}$, \label{lemma:bounds:sigma}

        \item $\| \bar{\mu}_{\theta,\ell}(x_i)_c \| \le U_{i,\ell}$. \label{lemma:bounds:mu_bar}

        \item $\| \bar{\sigma}_{\theta,\ell}(x_i)_c \| \le \sqrt{2} U_{i,\ell} + \sqrt{\eta}$, \label{lemma:bounds:sigma_bar}
    \end{enumerate}
    where  $U_{i,\ell}$ is provided in Assumption~\ref{assump:h}.
    
\end{lemma}

Note that
Lemma~\ref{lemma:bounds}-\ref{lemma:bounds:sigma} together with the definition given in Eq.~\eqref{eqn:mu_sig} establish the bounds
\begin{equation} \label{eqn:sigma_sandwich}
    \sqrt{\eta} \le \| \sigma(h_{\theta_\ell}(x_i))_{b,c} \| \le U_{i,\ell}.
\end{equation}

Equipped with Lemma~\ref{lemma:bounds}, we move on to presenting the next key result which establishes the Lipschitz continuity of all style statistics used in \textsc{StyleDDG}.

\section{Proof of Lemma~\ref{lemma:bounds}} \label{app:lemma:bounds}
\begin{proof}
    \ref{lemma:bounds:mu} Using the definition of $\mu(\cdot)_{b,c}$ from Eq.~\eqref{eqn:mu_sig}, we have
    \begin{equation}
        \begin{aligned}
            & \| \mu(h_{\theta_\ell}(x_i))_{b,c} \| = \left\| \frac1{HW} \sum_{h=1}^H{\sum_{w=1}^W{h_{\theta_\ell}(x_i)_{b,c,h,w}}} \right\|
            \\
            & \le \frac1{HW} \sum_{h=1}^H{\sum_{w=1}^W{\| h_{\theta_\ell}(x_i)_{b,c,h,w} \|}} \le \frac1{HW} \sum_{h=1}^H{\sum_{w=1}^W{U_{i,\ell}}}
            \\
            & = U_{i,\ell},
        \end{aligned}
    \end{equation}
    in which we used the triangle inequality.
\end{proof}

\begin{proof}
    \ref{lemma:bounds:sigma} We first find an upper bound for $\| \sigma^2(h_{\theta_\ell}(x_i))_{b,c} \|$, and then take the square root of those to conclude the proof. Using the definition of $\sigma^2(\cdot)_{b,c}$ from Eq.~\eqref{eqn:mu_sig}, we first rewrite its definition as
    \begin{equation}
        \begin{aligned}
            & \sigma^2(h_{\theta_\ell}(x_i))_{b,c}
            \\
            & = \frac1{HW} \sum_{h=1}^H{\sum_{w=1}^W{{\left( h_{\theta_\ell}(x_i)_{b,c,h,w} - \mu(h_{\theta_\ell}(x_i))_{b,c} \right)}^2}} + \eta
            \\
            & = \frac1{HW} \sum_{h=1}^H{\sum_{w=1}^W{{ h_{\theta_\ell}(x_i)_{b,c,h,w}^2}}} - \mu(h_{\theta_\ell}(x_i))_{b,c}^2 + \eta.
        \end{aligned}
    \end{equation}
    Using this definition, we can write the following
    \begin{equation}
        \begin{aligned}
            & \| \sigma^2(h_{\theta_\ell}(x_i))_{b,c} \|
            \\
            & = \left\| \frac1{HW} \sum_{h=1}^H{\sum_{w=1}^W{{ h_{\theta_\ell}(x_i)_{b,c,h,w}^2}}} - \mu(h_{\theta_\ell}(x_i))_{b,c}^2 + \eta \right\|
            \\
            & \le \frac1{HW} \sum_{h=1}^H{\sum_{w=1}^W{\| h_{\theta_\ell}(x_i)_{b,c,h,w}^2 \|}} + \| \mu(h_{\theta_\ell}(x_i))_{b,c}^2 \| + \eta \hspace{-3mm}
            \\
            & \le \frac1{HW} \sum_{h=1}^H{\sum_{w=1}^W{U_{i,\ell}^2}} + U_{i,\ell}^2 + \eta = 2U_{i,\ell}^2 + \eta,
        \end{aligned}
    \end{equation}
    in which we used the triangle inequality, and then invoked the results of Assumption~\ref{assump:h}-\ref{assump:h:bounded} and Lemma~\ref{lemma:bounds}-\ref{lemma:bounds:mu}. Taking the square root of the above inequality, we get
    \begin{equation}
        \| \sigma(h_{\theta_\ell}(x_i))_{b,c} \| \le \sqrt{2U_{i,\ell}^2 + \eta} \le \sqrt{2} U_{i,\ell} + \sqrt{\eta}.
    \end{equation}
\end{proof}

\begin{proof} \ref{lemma:bounds:mu_bar}
    Using the definition of $\bar{\mu}_{\theta,\ell}(x_i)_c$ from Eq.~\eqref{eqn:style_statistics}, we have
    \begin{equation}
        \begin{aligned}
            & \| \bar{\mu}_{\theta,\ell}(x_i)_c \| = \left\| \frac1B \sum_{b=1}^B{\mu(h_{\theta_\ell}(x_i))_{b,c}} \right\|
            \\
            & \le \frac1B \sum_{b=1}^B{\| \mu(h_{\theta_\ell}(x_i))_{b,c} \|} \le \frac1B \sum_{b=1}^B{U_{i,\ell}} = U_{i,\ell},
        \end{aligned}
    \end{equation}
    in which we used the triangle inequality and then invoked the result of Lemma~\ref{lemma:bounds}-\ref{lemma:bounds:mu}.
\end{proof}

\begin{proof}
    \ref{lemma:bounds:sigma_bar} Using the definition of $\bar{\sigma}_{\theta,\ell}(x_i)_c$ from Eq.~\eqref{eqn:style_statistics}, we have
    \begin{equation}
        \begin{aligned}
            \| \bar{\sigma}_{\theta,\ell}(x_i)_c \| & = \left\| \frac1B \sum_{b=1}^B{\sigma(h_{\theta_\ell}(x_i))_{b,c}} \right\|
            \\
            &  \le \frac1B \sum_{b=1}^B{\| \sigma(h_{\theta_\ell}(x_i))_{b,c} \|} \le \frac1B \sum_{b=1}^B{(\sqrt{2} U_{i,\ell} + \sqrt{\eta})}
            \\
            & = \sqrt{2} U_{i,\ell} + \sqrt{\eta},
        \end{aligned}
    \end{equation}
    in which we used the triangle inequality and then invoked the result of Lemma~\ref{lemma:bounds}-\ref{lemma:bounds:sigma}.
\end{proof}

\section{Proof of Proposition~\ref{proposition:lipschitz}} \label{app:proposition:lipschitz}

\begin{proof}
    \ref{proposition:lipschitz:mu} Using the definition $\bar{\mu}_{\theta,\ell}(x_i)_c = \frac1B \sum_{b=1}^B{\mu(h_{\theta_\ell}(x_i))_{b,c}}$ given in Eq.~\eqref{eqn:style_statistics}, we have that
    \begin{equation} \label{eqn:mu_bound}
        \begin{aligned}
            & \| \bar{\mu}_{\theta,\ell}(x_i)_c - \bar{\mu}_{\theta',\ell}(x_i)_c \|
            \\
            & \le \left\| \frac1B \sum_{b=1}^B{( \mu(h_{\theta_\ell}(x_i))_{b,c} - \mu(h_{\theta'_\ell}(x_i))_{b,c})} \right\|
            \\
            & \le \frac1B \sum_{b=1}^B{\| \mu(h_{\theta_\ell}(x_i))_{b,c} - \mu(h_{\theta'_\ell}(x_i))_{b,c} \|},
        \end{aligned}
    \end{equation}
    in which we used the triangle inequality. Therefore, we must establish Lipschitz continuity of $\mu(h_{\theta_\ell}(x_i)_{b,c})$. Using the definition of the function $\mu(\cdot)_{b,c}$ given in Eq.~\eqref{eqn:mu_sig}, we have that
    \begin{equation} \label{eqn:mu_bound_instance}
        \begin{aligned}
            & \| \mu(h_{\theta_\ell}(x_i))_{b,c} - \mu(h_{\theta'_\ell}(x_i))_{b,c} \|
            \\
            & \le \left\| \frac1{HW} \sum_{h=1}^H{\sum_{w=1}^W{(h_{\theta_\ell}(x_i)_{b,c,h,w} - h_{\theta'_\ell}(x_i)_{b,c,h,w})}} \right\|
            \\
            & \le \frac1{HW} \sum_{h=1}^H{\sum_{w=1}^W{\| h_{\theta_\ell}(x_i)_{b,c,h,w} - h_{\theta'_\ell}(x_i)_{b,c,h,w} \|}}
            \\
            & \le \frac1{HW} \sum_{h=1}^H{\sum_{w=1}^W{D_{i,\ell} \| \theta - \theta' \|}} = D_{i,\ell} \| \theta - \theta' \|,
        \end{aligned}
    \end{equation}
    where we used the triangle inequality first, then invoked Assumption~\ref{assump:h}-\ref{assump:h:lipschitz}. Plugging the result of Eq.~\eqref{eqn:mu_bound_instance} back in Eq.~\eqref{eqn:mu_bound}, we obtain
    \begin{equation}
        \| \bar{\mu}_{\theta,\ell}(x_i)_c - \bar{\mu}_{\theta',\ell}(x_i)_c \| \le \frac1B \sum_{b=1}^B{D_{i,\ell} \| \theta - \theta' \|} = D_{i,\ell} \| \theta - \theta' \|.
    \end{equation}
\end{proof}

\begin{proof}
    \ref{proposition:lipschitz:sigma} Using the definition $\bar{\sigma}_{\theta,\ell}(x_i)_c = \frac1B \sum_{b=1}^B{\sigma(h_{\theta_\ell}(x_i))_{b,c}}$ given in Eq.~\eqref{eqn:style_statistics}, we have that
    \begin{equation} \label{eqn:sigma_bound}
        \begin{aligned}
            & \| \bar{\sigma}_{\theta,\ell}(x_i)_c - \bar{\sigma}_{\theta',\ell}(x_i)_c \|
            \\
            & = \left\| \frac1B \sum_{b=1}^B{(\sigma(h_{\theta_\ell}(x_i))_{b,c} - \sigma(h_{\theta'_\ell}(x_i))_{b,c})} \right\|
            \\
            & \le \frac1B \sum_{b=1}^B{\| \sigma(h_{\theta_\ell}(x_i))_{b,c} - \sigma(h_{\theta'_\ell}(x_i))_{b,c} \|},
        \end{aligned}
    \end{equation}
    in which we used the triangle inequality. Thus, we must establish Lipschitz continuity of $\sigma(h_{\theta_\ell}(x_i))_{b,c}$. Towards this, we first use the difference of squares rule to obtain
    \begin{equation} \label{eqn:sigma_bound_instance}
        \begin{aligned}
            & \| \sigma(h_{\theta_\ell}(x_i))_{b,c} - \sigma(h_{\theta'_\ell}(x_i))_{b,c} \|
            \\
            & = \frac{\| \sigma^2(h_{\theta_\ell}(x_i))_{b,c} - \sigma^2(h_{\theta'_\ell}(x_i))_{b,c} \|}{\| \sigma(h_{\theta_\ell}(x_i))_{b,c} + \sigma(h_{\theta'_\ell}(x_i))_{b,c} \|}
            \\
            & \geq \frac{\| \sigma^2(h_{\theta_\ell}(x_i))_{b,c} - \sigma^2(h_{\theta'_\ell}(x_i))_{b,c} \|}{\min_{\theta_\ell, x_i}\{ \| \sigma(h_{\theta_\ell}(x_i))_{b,c} \| \}}
            \\
            & \geq \frac1{\sqrt{\eta}} \| \sigma^2(h_{\theta_\ell}(x_i))_{b,c} - \sigma^2(h_{\theta'_\ell}(x_i))_{b,c} \|,
        \end{aligned}
    \end{equation}
    where we used the fact that $\sigma(\cdot)_{b,c} > 0$ is always non-negative, and then invoked the lower bound of Eq.~\eqref{eqn:sigma_sandwich}. Similar to before, we must establish Lipschitz continuity of $\sigma^2(h_{\theta_\ell}(x_i))_{b,c}$ to prove Eq.~\eqref{eqn:sigma_bound_instance} (and in turn of Eq.~\eqref{eqn:sigma_bound}). Using the definition of the function $\sigma^2(\cdot)_{b,c}$ in Eq.~\eqref{eqn:mu_sig}, we have that
    \begin{equation}
        \begin{aligned}
            & \sigma^2(h_{\theta_\ell}(x_i))_{b,c}
            \\
            & = \frac1{HW} \sum_{h=1}^H{\sum_{w=1}^W{{\left( h_{\theta_\ell}(x_i)_{b,c,h,w} - \mu(h_{\theta_\ell}(x_i))_{b,c} \right)}^2}} + \eta
            \\
            & = \frac1{HW} \sum_{h=1}^H{\sum_{w=1}^W{h_{\theta_\ell}(x_i)_{b,c,h,w}^2}} - \mu(h_{\theta_\ell}(x_i))_{b,c}^2 + \eta.
        \end{aligned}
    \end{equation}
    Thus, we can obtain the following for the last term in Eq.~\eqref{eqn:sigma_bound_instance}
    \begin{equation} \label{eqn:sigma_bound_instance_squared}
        \begin{aligned}
            & \| \sigma^2(h_{\theta_\ell}(x_i))_{b,c} - \sigma^2(h_{\theta'_\ell}(x_i))_{b,c} \|
            \\
            & \begin{aligned}
                = \Bigg\| \frac1{HW} \sum_{h=1}^H{\sum_{w=1}^W{\left( h_{\theta_\ell}(x_i)_{b,c,h,w}^2 - h_{\theta'_\ell}(x_i)_{b,c,h,w}^2 \right)}}
                \\
                - \left( \mu(h_{\theta_\ell}(x_i))_{b,c}^2 - \mu(h_{\theta'_\ell}(x_i))_{b,c}^2 \right) \Bigg\|
            \end{aligned}
            \\
            & \begin{aligned}
                \le \frac1{HW} & \sum_{h=1}^H{\sum_{w=1}^W{\| h_{\theta_\ell}(x_i)_{b,c,h,w}^2 - h_{\theta'_\ell}(x_i)_{b,c,h,w}^2 \| }}
                \\
                & + \| \mu(h_{\theta_\ell}(x_i))_{b,c}^2 - \mu(h_{\theta'_\ell}(x_i))_{b,c}^2 \|,
            \end{aligned}
        \end{aligned}
    \end{equation}
    in which we used the triangle inequality. To bound Eq.~\eqref{eqn:sigma_bound_instance_squared}, we focus on its constituents. We first have that
    \begin{equation} \label{eqn:sigma_bound_instance_squared:first}
        \begin{aligned}
            & \| h_{\theta_\ell}(x_i)_{b,c,h,w}^2 - h_{\theta'_\ell}(x_i)_{b,c,h,w}^2 \|
            \\
            & \begin{aligned}
                = \| h_{\theta_\ell}(x_i)_{b,c,h,w} + & h_{\theta'_\ell}(x_i)_{b,c,h,w} \|
                \\
                & \| h_{\theta_\ell}(x_i)_{b,c,h,w} - h_{\theta'_\ell}(x_i)_{b,c,h,w} \|
            \end{aligned}
            \\
            & \le 2 \max_{\theta_\ell}\{\| h_{\theta_\ell}(x_i)_{b,c,h,w} \|\} D_{i,\ell} \| \theta - \theta' \|
            \\
            & = 2 U_i D_{i,\ell} \| \theta - \theta' \|,
        \end{aligned}
    \end{equation}
    where we used the difference of squares formula, and then invoked Assumption~\ref{assump:h}-\ref{assump:h:lipschitz} and \ref{assump:h}-\ref{assump:h:bounded}. For the second term in Eq.~\eqref{eqn:sigma_bound_instance_squared}, we have
    \begin{equation} \label{eqn:sigma_bound_instance_squared:second}
        \begin{aligned}
            & \| \mu(h_{\theta_\ell}(x_i))_{b,c}^2 - \mu(h_{\theta'_\ell}(x_i))_{b,c}^2 \|
            \\
            & \begin{aligned}
                = \| \mu(h_{\theta_\ell}(x_i))_{b,c} + \mu&(h_{\theta'_\ell}(x_i))_{b,c} \|
                \\
                & \| \mu(h_{\theta_\ell}(x_i))_{b,c} - \mu(h_{\theta'_\ell}(x_i))_{b,c} \|
            \end{aligned}
            \\
            & \le 2 \max_{\theta_\ell}\{\| \mu(h_{\theta_\ell}(x_i))_{b,c} \|\} D_{i,\ell} \| \theta - \theta' \|
            \\
            & = 2 U_i D_{i,\ell} \| \theta - \theta' \|,
        \end{aligned}
    \end{equation}
    where we used the difference of squares formula, and then invoked the results of Lemma~\ref{lemma:bounds}-\ref{lemma:bounds:mu} and Proposition~\ref{proposition:lipschitz}-\ref{proposition:lipschitz:mu}. Putting Eqs.~\eqref{eqn:sigma_bound_instance_squared:first} and \eqref{eqn:sigma_bound_instance_squared:second} together and plugging them back in \eqref{eqn:sigma_bound_instance_squared}, and then we get
    \begin{equation} \label{eqn:sigma_bound_instance_squared_final}
        \begin{aligned}
            & \| \sigma^2(h_{\theta_\ell}(x_i))_{b,c} - \sigma^2(h_{\theta'_\ell}(x_i))_{b,c} \|
            \\
            & \le \frac1{HW} \sum_{h=1}^H{\sum_{w=1}^W{2 U_{i,\ell} D_{i,\ell} \| \theta - \theta' \|}} + 2 U_i D_{i,\ell} \| \theta - \theta' \|
            \\
            & = 4 U_i D_{i,\ell} \| \theta - \theta' \|.
        \end{aligned}
    \end{equation}
    Finally, we employ Eq.~\eqref{eqn:sigma_bound_instance_squared_final} in Eq.~\eqref{eqn:sigma_bound_instance} followed by Eq.~\eqref{eqn:sigma_bound} to get
    \begin{equation}
        \begin{aligned}
            \| \bar{\sigma}_{\theta,\ell}(x_i)_c - \bar{\sigma}_{\theta',\ell}(x_i)_c \| & \le \frac1B \sum_{b=1}^B{\frac{4 U_i D_{i,\ell}}{\sqrt{\eta}} \| \theta - \theta' \|}
            \\
            & = \frac{4 U_i D_{i,\ell}}{\sqrt{\eta}} \| \theta - \theta' \|.
        \end{aligned}
    \end{equation}
\end{proof}

\begin{proof}
    \ref{proposition:lipschitz:Sigma_mu} Using the definition $\Sigma_{\mu,\theta,\ell}^2(x_i)_c = \Sigma_\mu^2(h_{\theta_\ell}(x_i))_c$ given in Eq.~\eqref{eqn:style_statistics} and the definition of the function $\Sigma_{\mu}^2(\cdot)_c$ in Eq.~\eqref{eqn:Sigma_mu_Sigma_sig}, we have that
    \begin{equation}
        \begin{aligned}
            & \Sigma_{\mu,\theta,\ell}^2(x_i)_c
            \\
            & = \frac1B \sum_{b=1}^B{{\left( \mu(h_{\theta_\ell}(x_i))_{b,c} - \mathbb{E}_b{[\mu(h_{\theta_\ell}(x_i))_{b,c}]} \right)}^2} + \eta
            \\
            & = \frac1B \sum_{b=1}^B{ \mu(h_{\theta_\ell}(x_i))_{b,c}^2} - \bar{\mu}_{\theta,\ell}(x_i)_c^2 + \eta,
        \end{aligned}
    \end{equation}
    where we used the definition of $\bar{\mu}_{\theta,\ell}(x_i)_c$ from Eq.~\eqref{eqn:style_statistics}. Consequently, we can obtain the bound
    \begin{equation} \label{eqn:Sigma_mu_bound}
        \begin{aligned}
            & \| \Sigma_{\mu,\theta,\ell}^2(x_i)_c - \Sigma_{\mu,\theta',\ell}^2(x_i)_c \|
            \\
            & \begin{aligned}
                = \Bigg\| \frac1B & \sum_{b=1}^B{\left( \mu(h_{\theta_\ell}(x_i))_{b,c}^2 - \mu(h_{\theta'_\ell}(x_i))_{b,c}^2 \right)}
                \\
                & - (\bar{\mu}_{\theta,\ell}(x_i)_c^2 - \bar{\mu}_{\theta',\ell}(x_i)_c^2) \Bigg\|
            \end{aligned}
            \\
            & \begin{aligned}
                \le \frac1B & \sum_{b=1}^B{\| \mu(h_{\theta_\ell}(x_i))_{b,c}^2 - \mu(h_{\theta'_\ell}(x_i))_{b,c}^2 \|}
                \\
                & + \| \bar{\mu}_{\theta,\ell}(x_i)_c^2 - \bar{\mu}_{\theta',\ell}(x_i)_c^2 \|,
            \end{aligned}
        \end{aligned}
    \end{equation}
    in which we used the triangle inequality. We have obtained a bound for the first term of Eq.~\eqref{eqn:Sigma_mu_bound} in Eq.~\eqref{eqn:sigma_bound_instance_squared:second}, and we focus on the second term next. We have that
    \begin{equation} \label{eqn:mu_bound_squared}
        \begin{aligned}
            & \| \bar{\mu}_{\theta,\ell}(x_i)_c^2 - \bar{\mu}_{\theta',\ell}(x_i)_c^2 \|
            \\
            & = \| \bar{\mu}_{\theta,\ell}(x_i)_c + \bar{\mu}_{\theta',\ell}(x_i)_c \| \| \bar{\mu}_{\theta,\ell}(x_i)_c - \bar{\mu}_{\theta',\ell}(x_i)_c \|
            \\
            & \le 2 \max_{\theta, \ell}\{ \| \bar{\mu}_{\theta,\ell}(x_i)_c \| \} D_{i,\ell} \| \theta - \theta' \|
            \\
            & = 2 U_i D_{i,\ell} \| \theta - \theta' \|,
        \end{aligned}
    \end{equation}
    where we used the difference of squares law, and then invoke the results of Lemma~\ref{lemma:bounds}-\ref{lemma:bounds:mu_bar} and Proposition~\ref{proposition:lipschitz}-\ref{proposition:lipschitz:mu}, respectively. To conclude the proof, we plug back Eqs.~\eqref{eqn:mu_bound_squared} and \eqref{eqn:sigma_bound_instance_squared:second} back in Eq.~\eqref{eqn:Sigma_mu_bound} to get
    \begin{equation}
        \begin{aligned}
            & \| \Sigma_{\mu,\theta,\ell}^2(x_i)_c - \Sigma_{\mu,\theta',\ell}^2(x_i)_c \|
            \\
            & \le \frac1B \sum_{b=1}^B{2 U_i D_{i,\ell} \| \theta - \theta' \|} + 2 U_i D_{i,\ell} \| \theta - \theta' \|
            \\
            & = 4 U_i D_{i,\ell} \| \theta - \theta' \|.
        \end{aligned}
    \end{equation}
\end{proof}

\begin{proof}
    \ref{proposition:lipschitz:Sigma_sigma} Using the definition $\Sigma_{\sigma,\theta,\ell}^2(x_i)_c = \Sigma_\sigma^2(h_{\theta_\ell}(x_i))_c$ given in Eq.~\eqref{eqn:style_statistics} and the definition of the function $\Sigma_{\sigma}^2(\cdot)_c$ in Eq.~\eqref{eqn:Sigma_mu_Sigma_sig}, we have that
    \begin{equation}
        \begin{aligned}
            & \Sigma_{\sigma,\theta,\ell}^2(x_i)_c
            \\
            & = \frac1B \sum_{b=1}^B{{\left( \sigma(h_{\theta_\ell}(x_i))_{b,c} - \mathbb{E}_b{[\sigma(h_{\theta_\ell}(x_i))_{b,c}]} \right)}^2} + \eta
            \\
            & = \frac1B \sum_{b=1}^B{ \sigma(h_{\theta_\ell}(x_i))_{b,c}^2} - \bar{\sigma}_{\theta,\ell}^2(x_i)_c + \eta,
        \end{aligned}
    \end{equation}
    where we used the definition of $\bar{\sigma}_{\theta,\ell}^2(x_i)_c$ from Eq.~\eqref{eqn:style_statistics}. Hence, we can obtain the bound
    \begin{equation} \label{eqn:Sigma_sigma_bound}
        \begin{aligned}
            & \| \Sigma_{\sigma,\theta,\ell}^2(x_i)_c - \Sigma_{\sigma,\theta',\ell}^2(x_i)_c \|
            \\
            & \begin{aligned}
                = \Bigg\| \frac1B & \sum_{b=1}^B{\left( \sigma(h_{\theta_\ell}(x_i))_{b,c}^2 - \sigma(h_{\theta'_\ell}(x_i))_{b,c}^2 \right)}
                \\
                & - (\bar{\sigma}_{\theta,\ell}^2(x_i)_c - \bar{\sigma}_{\theta',\ell}^2(x_i)_c) \Bigg\|
            \end{aligned}
            \\
            & \begin{aligned}
                \le \frac1B & \sum_{b=1}^B{\| \sigma(h_{\theta_\ell}(x_i))_{b,c}^2 - \sigma(h_{\theta'_\ell}(x_i))_{b,c}^2 \|}
                \\
                & + \| \bar{\sigma}_{\theta,\ell}^2(x_i)_c - \bar{\sigma}_{\theta',\ell}^2(x_i)_c\|,
            \end{aligned}
        \end{aligned}
    \end{equation}
    in which we used the triangle inequality. We have obtained bounds for the first term of Eq.~\eqref{eqn:Sigma_sigma_bound} in Eq.~\eqref{eqn:sigma_bound_instance_squared_final}, and we focus on the second term next. We have that
    \begin{equation} \label{eqn:sigma_bound_squared}
        \begin{aligned}
            & \| \bar{\sigma}_{\theta,\ell}(x_i)_c^2 - \bar{\sigma}_{\theta',\ell}(x_i)_c^2 \|
            \\
            & = \| \bar{\sigma}_{\theta,\ell}(x_i)_c + \bar{\sigma}_{\theta',\ell}(x_i)_c \| \| \bar{\sigma}_{\theta,\ell}(x_i)_c - \bar{\sigma}_{\theta',\ell}(x_i)_c \|
            \\
            & \le 2 \max_{\theta,\ell}\{ \| \bar{\sigma}_{\theta,\ell}(x_i)_c \| \} \frac{4 U_i D_{i,\ell}}{\sqrt{\eta}} \| \theta - \theta' \|
            \\
            & = \frac{8 \sqrt{2} U_i^2 D_{i,\ell}}{\sqrt{\eta}} \| \theta - \theta' \|,
        \end{aligned}
    \end{equation}
    where we used the difference of squares law, and then invoke the results of Lemma~\ref{lemma:bounds}-\ref{lemma:bounds:sigma_bar} and Proposition~\ref{proposition:lipschitz}-\ref{proposition:lipschitz:sigma}, respectively. To conclude the proof, we plug back Eqs.~\eqref{eqn:sigma_bound_squared} and \eqref{eqn:sigma_bound_instance_squared_final} back in Eq.~\eqref{eqn:Sigma_mu_bound} to get
    \begin{equation}
        \begin{aligned}
            & \| \Sigma_{\sigma,\theta,\ell}^2(x_i)_c - \Sigma_{\sigma,\theta',\ell}^2(x_i)_c \|
            \\
            & \le \frac1B \sum_{b=1}^B{4 U_i D_{i,\ell} \| \theta - \theta' \|} + \frac{8 \sqrt{2} U_i^2 D_{i,\ell}}{\sqrt{\eta}} \| \theta - \theta' \|
            \\
            & = 4 U_i D_{i,\ell} \left( 1 + \frac{2\sqrt{2} U_i}{\sqrt{\eta}} \right) \| \theta - \theta' \|.
        \end{aligned}
    \end{equation}
\end{proof}

\section{Proof of Proposition~\ref{proposition:smooth}} \label{app:proposition:smooth}

\begin{proof}
    With the bounds for all $4$ individual style statistics of each device $i \in \mathcal{M}$ at a given layer/block $\ell = \{ 1, ..., L \}$ derived in Proposition~\ref{proposition:lipschitz}, we move on to analyze the Lipschitz continuity of $\psi_{\theta,\ell}(x_i) = \begin{bmatrix} \bar{\mu}_{\theta,\ell}(x_i) & \bar{\sigma}_{\theta,\ell}(x_i) & \Sigma_{\mu,\theta,\ell}^2(x_i) & \Sigma_{\sigma,\theta,\ell}^2(x_i) \end{bmatrix}^T$, which is a vector concatenating all of them for a single client $i$ at a single layer $\ell$. We have
    \begin{equation}
        \begin{aligned}
            & \| \psi_{\theta,\ell}(x_i) - \psi_{\theta',\ell}(x_i) \|
            \\
            & \begin{aligned}
                \le \| & \bar{\mu}_{\theta,\ell}(x_i) - \bar{\mu}_{\theta',\ell}(x_i) \| + \| \bar{\sigma}_{\theta,\ell}(x_i) - \bar{\sigma}_{\theta',\ell}(x_i) \|
                \\
                & + \| \Sigma_{\mu,\theta,\ell}^2(x_i) - \Sigma_{\mu,\theta',\ell}^2(x_i) \| + \| \Sigma_{\sigma,\theta,\ell}^2(x_i) - \Sigma_{\sigma,\theta',\ell}^2(x_i) \|
            \end{aligned}
            \\
            & \resizebox{\linewidth}{!}{$\le \left( D_{i,\ell} + \frac{4 U_i D_{i,\ell}}{\sqrt{\eta}} + 4 U_i D_{i,\ell} + 4 U_i D_{i,\ell} \left( 1 + \frac{2\sqrt{2} U_i}{\sqrt{\eta}} \right) \right) \| \theta - \theta' \|$}
            \\
            & = \left( 1 + 4U_i \left( 1/\sqrt{\eta} + 2 + (2\sqrt{2} U_i) / \sqrt{\eta} \right) \right) D_{i,\ell} \| \theta - \theta' \|,
        \end{aligned}
    \end{equation}
    in which we used the triangle inequality. Next, we focus on $\psi_\theta(x_i) = \begin{bmatrix} \psi_{\theta,1}(x_i) & \cdots & \psi_{\theta,L}(x_i) \end{bmatrix}^T$, which is a vector concatenating the style statistics across all the layers of the CNN for a given client $i \in \mathcal{M}$. We have that
    \begin{equation}
        \begin{aligned}
            & \| \psi_\theta(x_i) - \psi_{\theta'}(x_i) \| \le \sum_{\ell=1}^L{\| \psi_{\theta,\ell}(x_i) - \psi_{\theta',\ell}(x_i) \|}
            \\
            & \le L \left( 1 + 4U_i \left( 1/\sqrt{\eta} + 2 + (2\sqrt{2} U_i) / \sqrt{\eta} \right) \right) D_i \| \theta - \theta' \|, \hspace{-5mm}
        \end{aligned}
    \end{equation}
    in which we used the triangle inequality. Finally, the Lipschitzness of $\Psi_\theta(X) = \begin{bmatrix} \psi_\theta(x_1) & \cdots & \psi_\theta(x_m) \end{bmatrix}^T$ follows similarly as
    \begin{equation}
        \begin{aligned}
            & \| \Psi_\theta(X) - \Psi_{\theta'}(X) \| \le \sum_{i=1}^m{\| \psi_\theta(x_i) - \psi_{\theta'}(x_i) \|}
            \\
            & \le mL \left( 1 + 4U \left( 1/\sqrt{\eta} + 2 + (2\sqrt{2} U) / \sqrt{\eta} \right) \right) D \| \theta - \theta' \|, \hspace{-7mm}
        \end{aligned}
    \end{equation}
    in which we used the triangle inequality, and  $m$ is the total number of devices in the decentralized network. Noting that $\Psi_{\theta_i}(X) \subset \Psi_\theta(X)$, we thus have that
    \begin{equation}
        \begin{aligned}
            & \| \Psi_{\theta,i}(X) - \Psi_{\theta',i}(X) \|
            \\
            & \le mL \left( 1 + 4U \left( 1/\sqrt{\eta} + 2 + (2\sqrt{2} U) / \sqrt{\eta} \right) \right) D \| \theta - \theta' \|.
        \end{aligned}
    \end{equation}
    Plugging the last equation above into Assumption~\ref{assump:smooth}, we obtain
    \begin{equation}
        \begin{aligned}
            & \| \nabla{F}_i(\theta, \Psi_{\theta,i}(X)) - \nabla{F}_i(\theta', \Psi_{\theta',i}(X)) \|
            \\
            & \le \beta_{\theta,i} \| \theta - \theta' \| + \beta_{\Psi,i} \| \Psi_{\theta,i}(X)) - \Psi_{\theta',i}(X)) \|
            \\
            & \begin{aligned}
                \le \Big[ \beta_{\theta,i} + mL \Big( 1 + 4U & \left( 1/\sqrt{\eta} + 2 + (2\sqrt{2} U) / \sqrt{\eta} \right) \Big)
                \\
                & D \beta_{\Psi,i} \Big] \| \theta - \theta' \|.
            \end{aligned}
        \end{aligned}
    \end{equation}
\end{proof}


\end{document}